\documentclass[10pt,twocolumn]{article}
\usepackage[margin=0.75in,top=1in]{geometry}
\usepackage{cite}
\usepackage{amsmath,amssymb,amsfonts,mathtools}
\usepackage{textcomp}
\usepackage{booktabs}
\usepackage{cite}
\usepackage{graphicx}
\usepackage{textcomp}
\usepackage{multirow}
\usepackage{tablefootnote}
\usepackage[dvipsnames]{xcolor}
\usepackage{multirow}
\usepackage{cuted}
\usepackage{capt-of}
\usepackage{pifont}
\usepackage{fancyhdr,graphicx}
\usepackage[linesnumbered,ruled,vlined]{algorithm2e}
\usepackage{bm}
\def\BibTeX{{\rm B\kern-.05em{\sc i\kern-.025em b}\kern-.08em
    T\kern-.1667em\lower.7ex\hbox{E}\kern-.125emX}}

\begin{document}

\title{Active Diffusion Matching: Score-based Iterative Alignment of Cross-Modal Retinal Images}

\author{
Kanggeon Lee$^{1}$, Su Jeong Song$^{2}$, Soochahn Lee$^{3}$\thanks{Corresponding authors}, Kyoung Mu Lee$^{1}$\footnotemark[1] \\
$^{1}$ASRI, Dept. of ECE, Seoul National University, Korea \\
$^{2}$Dept. of Ophthalmology, Kangbuk Samsung Hospital, Sungkyunkwan University, Korea \\
$^{3}$School of Electrical Engineering, Kookmin University, Korea \\
\footnotesize\texttt{dlrkdrjs97@snu.ac.kr, sjsong7@gmail.com, sclee@kookmin.ac.kr, kyoungmu@snu.ac.kr}
}

\date{}
\maketitle

\begin{abstract}
\textit{Objective}:
The study aims to address the challenge of aligning Standard Fundus Images (SFIs) and Ultra-Widefield Fundus Images (UWFIs), which is difficult due to their substantial differences in viewing range and the amorphous appearance of the retina. 
Currently, no specialized method exists for this task, and existing image alignment techniques lack accuracy.

\textit{Methods}:
We propose Active Diffusion Matching (\textsc{ADM}), a novel cross-modal alignment method. 
\textsc{ADM} integrates two interdependent score-based diffusion models to jointly estimate global transformations and local deformations via an iterative Langevin Markov chain. 
This approach facilitates a stochastic, progressive search for optimal alignment. 
Additionally, custom sampling strategies are introduced to enhance the adaptability of \textsc{ADM} to given input image pairs.

\textit{Results}:
Comparative experimental evaluations demonstrate that \textsc{ADM} achieves state-of-the-art alignment accuracy. 
This was validated on two datasets: a private dataset of SFI-UWFI pairs and a public dataset of SFI-SFI pairs, 
with mAUC improvements of 5.2 and 0.4 points on the private and public datasets, respectively, compared to existing state-of-the-art methods.

\textit{Conclusion}:
\textsc{ADM} effectively bridges the gap in aligning SFIs and UWFIs, providing an innovative solution to a previously unaddressed challenge.
The method's ability to jointly optimize global and local alignment makes it highly effective for cross-modal image alignment tasks.

\textit{Significance}:
\textsc{ADM} has the potential to transform the integrated analysis of SFIs and UWFIs, enabling better clinical utility and supporting learning-based image enhancements. 
This advancement could significantly improve diagnostic accuracy and patient outcomes in ophthalmology.
\end{abstract}


\begin{figure*}[ht]
\includegraphics[width=\textwidth]{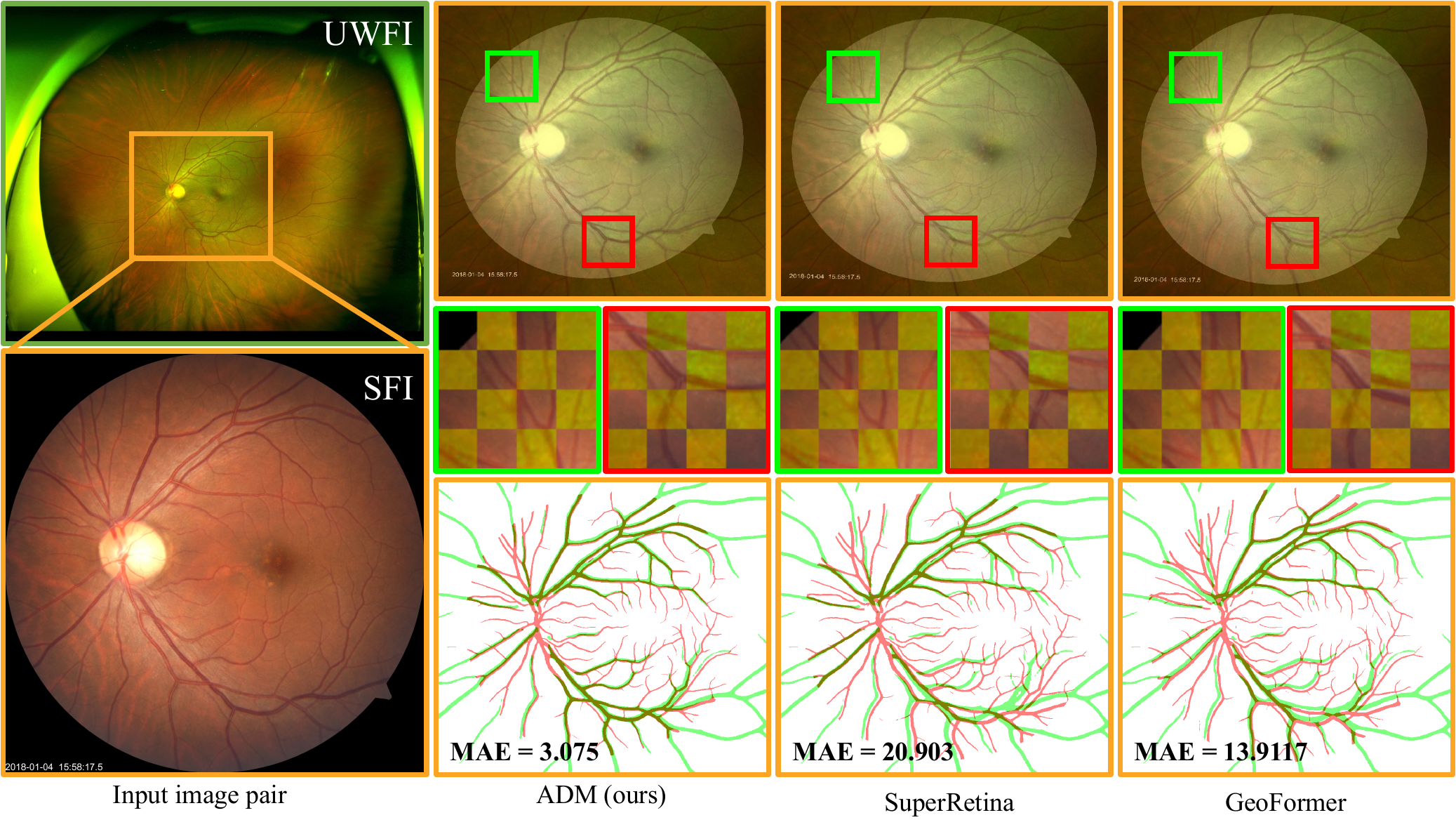}
\centering
\caption{
\textbf{Alignment of standard fundus images (SFIs) and ultra-widefield images (UWFIs) using \textsc{ADM}.}
We present a method for the alignment of SFI-UWFI pairs. 
The FOV of the SFI is limited to the orange box region of the UWFI. 
The cropped and zoomed-in green and red boxes highlight the alignment results of SuperRetina~\cite{liu2022semi}, GeoFormer~\cite{liu2023geometrized}, and our proposed \textsc{ADM}. 
The image below shows the intersection area between the SFI vessel (red line) and the UWFI vessel (green line), which exhibits the maximum alignment error (MAE)~\cite{wang2015robust, liu2022semi, truong2019glampoints, liu2023geometrized}.}
\label{fig:FIG_TEASER}
\end{figure*}

\section{Introduction}
\label{sec:intro}
Conventional standard fundus images (SFIs) typically capture only a central field of view ranging from 30$^{\circ}$ to 60$^{\circ}$, thereby covering less than 20$\%$ of the retinal area~\cite{lee2016ultra}. 
n contrast, ultra-widefield fundus images (UWFIs) enable visualization of up to 200$^{\circ}$ or approximately 82$\%$ of the retina within a single capture~\cite{lee2016ultra, witmer2013comparison}.
Consequently, UWFIs have become indispensable for the detection and assessment of retinal pathologies, such as diabetic retinopathy and retinal vascular occlusions, which predominantly affect the peripheral retina.

Although UWFIs significantly expand the field of view (FOV) and enhance diagnostic coverage, they compromise resolution and clarity relative to SFIs.
Consequently, UWFIs may prove inadequate for the detailed evaluation of critical retinal diseases that require close examination of retinal microstructures, such as age-related macular degeneration and diabetic retinopathy.

Therefore, there is considerable interest in enhancing the image quality of UWFIs through machine learning-based image enhancement~\cite{lee2023deep} and super-resolution techniques~\cite{lim2017enhanced}.
Achieving optimal performance with these methods necessitates large training datasets consisting of accurately aligned SFI-UWFI pairs, which in turn requires an automated and reliable SFI-UWFI alignment method.
To the best of our knowledge, no existing method has been specifically designed for this purpose.
Nonetheless, if precise alignment can be achieved, the quality of UWFIs could potentially be elevated to that of SFIs, thereby enabling UWFIs to fully supplant SFIs.

However, the alignment of SFI-UWFI pairs remains highly challenging due to substantial differences in field of view (FOV) and scale, as well as variations in color characteristics and the paucity of distinctive retinal textures.
Existing retinal image alignment methods~\cite{hernandez2020rempe,liu2022semi} have predominantly focused on aligning SFI-SFI pairs, which involve considerably smaller variations, especially in scale.
Current state-of-the-art image alignment approaches, such as those that estimate affine transformation parameters via single-step transformer inference~\cite{sun2021loftr,liu2023geometrized}, are insufficient to address the complex disparities present in SFI-UWFI pairs.
Moreover, iterative methods that determine local point correspondences often exhibit reduced accuracy when distinctive local feature points are sparse, as commonly observed in both SFIs and UWFIs.

Therefore, we propose a method to address the complex variations between SFI-UWFI pairs, as illustrated in Fig.~\ref{fig:FIG_TEASER}, by employing an iterative incremental alignment approach that gradually mitigates the extreme differences in scale and field of view.
At each iteration, a trained neural network progressively refines the alignment parameters for both the global transformation and local deformations, building upon previous estimates.
This refinement process is realized through a reverse diffusion process~\cite{ho2020denoising,chan2024tutorial}, driven by two interconnected score-based models~\cite{song2021scorebased} conditioned on the given input image pairs.
Each model iteratively produces refined estimates for the global transformation and local deformation, respectively, where feedback from the local deformation is utilized to correct inaccuracies in the global transformation.
The two models are trained end-to-end and function as the score function within Langevin dynamics~\cite{welling2011bayesian,song2019generative,song2021scorebased} during inference.

We term our method \emph{Active Diffusion Matching (\textsc{ADM})}, inspired by its similarity to the classic Active Shape Model (ASM)~\cite{cootes1995active}, which iteratively aligns a pre-trained shape model to a given image.
To the best of our knowledge, \textsc{ADM} is the first accurate and fully automatic method for aligning SFI-UWFI pairs, as prior works have only explored manually guided alignment~\cite{thuma2023big}.
\textsc{ADM} is a diffusion-based framework that effectively addresses the substantial global transformation and local deformations present in SFI-UWFI pairs, surpassing previous diffusion-based alignment methods that separately estimate local~\cite{kim2022diffusemorph} and global~\cite{wang2023posediffusion} variations.
Our quantitative evaluations demonstrate that \textsc{ADM} significantly outperforms state-of-the-art image alignment methods on a private dataset of SFI-UWFI pairs and achieves competitive performance on a public dataset of SFI-SFI pairs.

\begin{figure*}[ht!]
    \centering
    \includegraphics[width=1.0\textwidth]{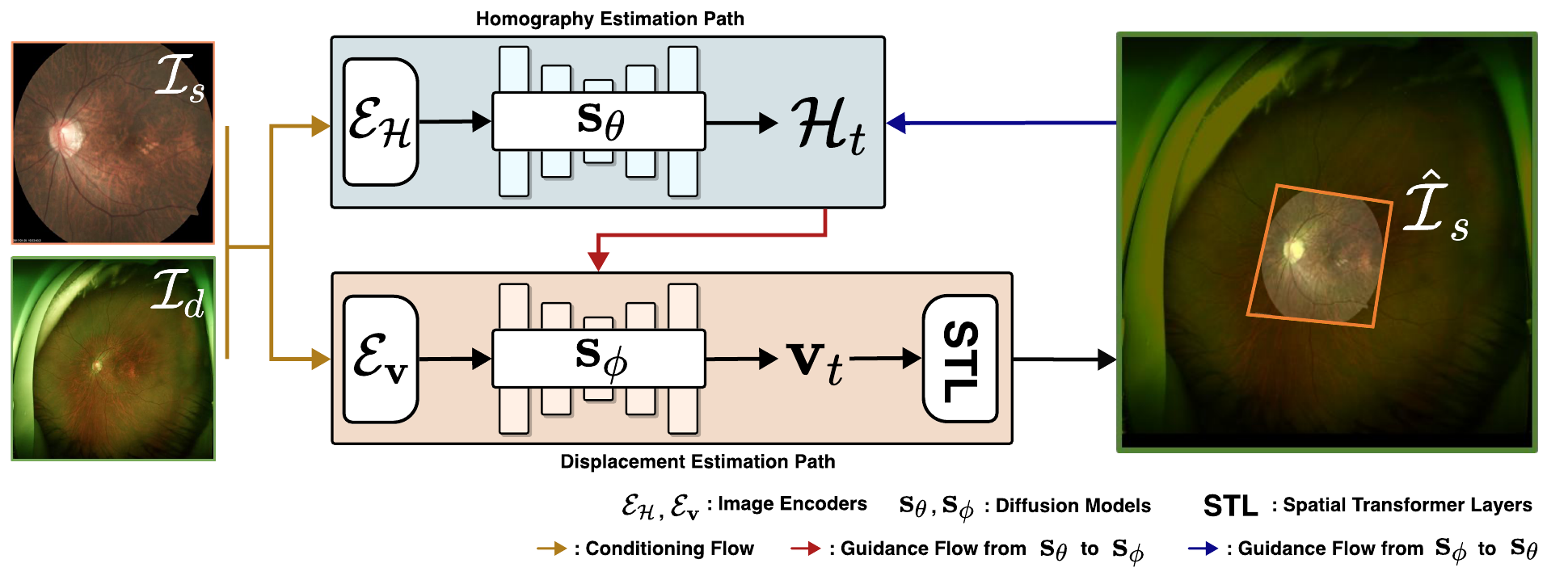}
    \caption{
    \textbf{Overview of \textsc{ADM}}. 
    \textsc{ADM} aligns the source image $\mathcal{I}_s$ (SFI) to the destination image $\mathcal{I}_d$ (UWFI) using a dual diffusion model architecture.
    Two score networks are employed: $\mathbf{s}_{\boldsymbol\theta}$ estimates global homography $\mathcal{H}$, while $\mathbf{s}_{\boldsymbol\phi}$ estimates local displacement field $\mathbf{v}$.
    Both networks are conditioned on the input image pair $(\mathcal{I}_s, \mathcal{I}_d)$ via dedicated encoders $\mathcal{E}_\mathcal{H}$ and $\mathcal{E}_{\mathbf{v}}$, which extract modality-adapted latent features.
    At each diffusion step $t$, $\mathbf{s}_{\boldsymbol\theta}$ estimates $\mathcal{H}_t$, and $\mathbf{s}_{\boldsymbol\phi}$ predicts $\mathbf{v}_t$ conditioned on both input images and the current estimate of $\mathcal{H}_t$.
    This cyclic interaction allows $\mathcal{H}_t$ to guide the estimation of $\mathbf{v}_t$, while the reverse influence is incorporated via a guidance term during the update of $\mathcal{H}_t$.
    The final aligned image $\hat{\mathcal{I}}_s$ is obtained by sequentially applying the predicted global transformation $\mathcal{H}$ and local deformation $\mathbf{v}$ to $\mathcal{I}_s$ through Spatial Transformer Layers (STL), adapted from~\cite{jaderberg2015spatial}.
    }
    \label{fig:FIG_ADM}
\end{figure*}

\section{Related Works}
\label{sec:related}
Here, we review related works on retinal image alignment and UWFI enhancement, as well as on general image alignment, including recent developments using diffusion models.

\noindent
{\bf{Retinal image alignment and UWFI enhancement.}}
While several methods have been proposed for registering SFIs with other imaging modalities~\cite{Nie2024}, such as optical coherence tomography (OCT)~\cite{lee2019adeep} or fluorescein angiography (FA)~\cite{noh2019fine}, we identified only one prior method addressing alignment with UWFIs~\cite{bioengineering11060568}, which relies on manual intervention.
Recently, a method for UWFI enhancement was introduced using unpaired learning to model the distinct characteristics of the SFI and UWFI datasets~\cite{lee2023deep}.

Several methods specifically designed for aligning SFI-SFI pairs have also been proposed.
REMPE~\cite{hernandez2020rempe} utilizes a 3D shape model of the eye to accommodate nonlinear deformations between image pairs.
The SuperRetina~\cite{liu2022semi} approach aligns SFI-SFI pairs by learning to detect and match retinal keypoints.
GeoFormer~\cite{liu2023geometrized} incorporates cross-attention layers to align potential common local regions.
Liu et al.~\cite{10821896} integrate a local alignment network into SuperRetina~\cite{liu2022semi} and GeoFormer~\cite{liu2023geometrized}, forming a two-step global-to-local alignment framework.
In contrast, \textsc{ADM} employs diffusion models that learn iterative global-local alignment to address the challenges of matching SFI-UWFI pairs, which involve differences not only in geometry but also in image domains.

\noindent
{\bf {General image alignment.}} 
Many methods have been proposed to determine the mapping between two planes in projective space by estimating a homography matrix~\cite{hartley2003multiple,szeliski2022computer}.
Compared to traditional keypoint-based methods relying on hand-crafted detectors and descriptors~\cite{lowe2004distinctive,bay2006surf,rosten2008faster,calonder2010brief}, recent machine learning approaches~\cite{detone2018superpoint,truong2019glampoints,revaud2019r2d2,sarlin2020superglue,liu2022semi,lindenberger2023lightglue} have demonstrated superior effectiveness.
Nevertheless, the scarcity of distinctive local regions may still limit the number of keypoints detected and thus constrain alignment accuracy.

Advances in neural network architectures have enabled detector-free direct regression methods~\cite{detone2016deep,rocco2020ncnet,sun2021loftr,zhang2022relpose,sinha2023sparsepose,wang2023posediffusion}.
While these methods exhibit flexibility to accommodate a wide range of transformations, they may still face challenges when dealing with extreme variations.

Many other methods employ iterative approaches for alignment.
Classic iterative frameworks such as Iterative Closest Points (ICP)~\cite{besl1992method} and Active Shape Models (ASM)~\cite{cootes1995active} perform well given good initializations.
More recent iterative estimation techniques~\cite{dong2018learning,zhao2021deep,cao2022iterative,cao2023recurrent,zhu2024mcnet,deng2024crosshomo} are capable of handling significant perspective warping, but their effectiveness diminishes when confronted with large-scale differences.

We also acknowledge methods for estimating local deformation, which generally assume that the image pairs are already reasonably well aligned at a global level.
These methods are typically applied in scenarios such as adjacent video frames for optical flow~\cite{zhang2021separable,xu2022gmflow,huang2022flowformer} or medical images of anatomical regions~\cite{cao2017deformable,hu2018weakly,xu2019deepatlas,balakrishnan2019voxelmorph,kim2021cyclemorph}.
Although these methods alone are unsuitable for image pairs exhibiting significant variations, they can be effectively combined with global alignment techniques to improve accuracy.
This approach is exemplified by the spatial transformer network~\cite{jaderberg2015spatial} and subsequent two-step global-local estimation methods~\cite{lee2019istn,de2019deep}.
A similar strategy is employed in \textsc{ADM}, but within an iterative incremental alignment framework.

\noindent
{\bf{Diffusion models for alignment.}} 
Diffusion models generate probabilistic data samples by simulating the reverse diffusion process, progressively transforming simple noise into complex data distributions through iterative refinement.
Although primarily employed for image synthesis~\cite{ho2020denoising,dhariwal2021diffusion,rombach2022high}, their effectiveness in estimation tasks has been demonstrated in methods for estimating local deformation fields~\cite{kim2022diffusemorph} and camera poses~\cite{wang2023posediffusion,zhang2024raydiffusion}.
However, no existing method has yet been proposed to jointly estimate both global and local alignment.

\section{Proposed Method}\label{sec:proposed_method}
\subsection{Score-based Langevin and Diffusion Models}
The Langevin dynamics~\cite{welling2011bayesian,song2019generative,song2021scorebased} for producing samples from a probability density $p(\mathbf{x})$ are defined as follows: 
\begin{equation}
    \mathbf{x}_{t+1} = \mathbf{x}_{t} + \epsilon_{t} \nabla_{\mathbf{x}} \log p(\mathbf{x}_{t}) + \sqrt{2\epsilon_{t}} \mathbf{z}_t,
    \label{eq:smld_samp}
\end{equation}
where $\mathbf{x}$ is the random variable representing the output parameters, $\epsilon_{t} \le 0$ is the step size, and $\mathbf{z}_t$ is noise sampled from the standard normal distribution~\cite{song2019generative}.
$\nabla_{\mathbf{x}} \log p(\mathbf{x}_{t})$, which is the gradient of $\log p(\mathbf{x}_{t})$, is defined as the score function of $p(\mathbf{x})$.
The addition of $\mathbf{z}_t$ converts Eq.~\ref{eq:smld_samp} from gradient descent to stochastic gradient descent~\cite{chan2024tutorial}, improving the robustness to gradient noise~\cite{scaman2020robustness}.

The score function can be trained as a noise conditional score network, denoted by $\mathbf {s}_{\mathbf \theta}(\mathbf{x}, \sigma) \approx \nabla_{\mathbf{x}} \log p(\mathbf{x}_{t})$, with respect to $\mathbf{x}$, using the denoising score matching objective function:
\begin{equation}
\resizebox{.9\hsize}{!}{$
    \boldsymbol{\theta}^{*} = \operatorname*{arg\,min}_{\boldsymbol{\theta}} \\ \sum_{t=1}^{N} 
    { 
    \sigma^{2}_{t} 
    \mathbb{E}_{p_{\mathcal{D}}(\mathbf{x})} 
    \mathbb{E}_{p_{\sigma_{t}}(\tilde{\mathbf{x}}|\mathbf{x})} 
    \left[
    \lVert
    \mathbf {s}_{\mathbf \theta}(\tilde{\mathbf{x}}, \sigma_{t}) - \nabla_{\tilde{\mathbf{x}}} \log p_{\sigma_{t}}(\tilde{\mathbf{x}}|\mathbf{x})
    \rVert^{2}_{2}
    \right]
    },
    $}
    \label{eq:smld_obj}
\end{equation}
with $p_{\sigma}(\tilde{\mathbf{x}}) \coloneq \int {p_{\mathcal{D}}(\mathbf{x})p_{\sigma_{t}}(\tilde{\mathbf{x}}|\mathbf{x})\text{d}\mathbf{x} }$ as the distribution of the noise-perturbed parameter $\tilde{\mathbf{x}}$, $p_{\sigma_{t}}(\tilde{\mathbf{x}}|\mathbf{x}) = \mathcal{N}(\tilde{\mathbf{x}};\mathbf{x}, \sigma^{2}_{t}\mathbf{I})$ 
as the Gaussian noise perturbation kernel, and $p_{\mathcal{D}}(\mathbf{x})$ as the data distribution. 
Eq.~\ref{eq:smld_samp} is then applied for inferring $\mathbf{x}$ values as:
\begin{equation}
    \mathbf{x}_{t+1} = \mathbf{x}_{t} + \epsilon_{t} \mathbf{s}_{\boldsymbol{\theta}^{\mathbf{*}}}(\mathbf{x}, \sigma_{t}) + \sqrt{2\epsilon_{t}} \mathbf{z}_t,
    \label{eq:smld_ddpm}
\end{equation}
with $t=1,2,\cdots,N$.

If we set the noise scales in Eq.~\ref{eq:smld_obj} to $\sigma^{2}_{t} = (1-\alpha_t)$ and define noise perturbation kernel as $p_{\alpha_{t}}({\mathbf{x}_{t}}|\mathbf{x}_{0}) = \mathcal{N}(\mathbf{x}_{t};\sqrt{\alpha_{t}}\mathbf{x}_{0}, (1-\alpha_{t})\mathbf{I})$, 
the objective score function $\mathbf {s}_{\mathbf \theta}(\mathbf {x}, t)$ is equivalent to that used in denoising diffusion probabilistic models~\cite{sohldickstein2015deep,ho2020denoising}. 
Accordingly, the Markov chain in the sampling process is modified to:
\begin{equation}
    \mathbf{x}_{t-1} = \frac{1}{\sqrt{1-\beta_{t}}}
    \mathbf{x}_{t} + \beta_{t} \mathbf{s}_{\boldsymbol{\theta}^{\mathbf{*}}}(\mathbf{x}, t) + \sqrt{\beta_{t}} \mathbf{z}_t,
    \label{eq:ddpm_samp}
\end{equation}
where $p_{\sigma_{t}}({\mathbf{x}}_{t}|\mathbf{x}_{t-1}) = \mathcal{N}({\mathbf{x}}_{t};\sqrt{1-\beta_{t}}\mathbf{x}_{t-1}, \beta_{t}\mathbf{I})$ is the noise kernel for a single iteration with variance $\beta_{t}$ and $t=N,N-1,\cdots,1$.
The noise scales $\alpha_{t}$ and $\beta_{t}$ are correlated as $\alpha_{t} = \prod^{t}_{j=1}(1-\beta_{j})$.

\begin{figure*}[h!]
    \centering
    \includegraphics[width=1.0\textwidth]{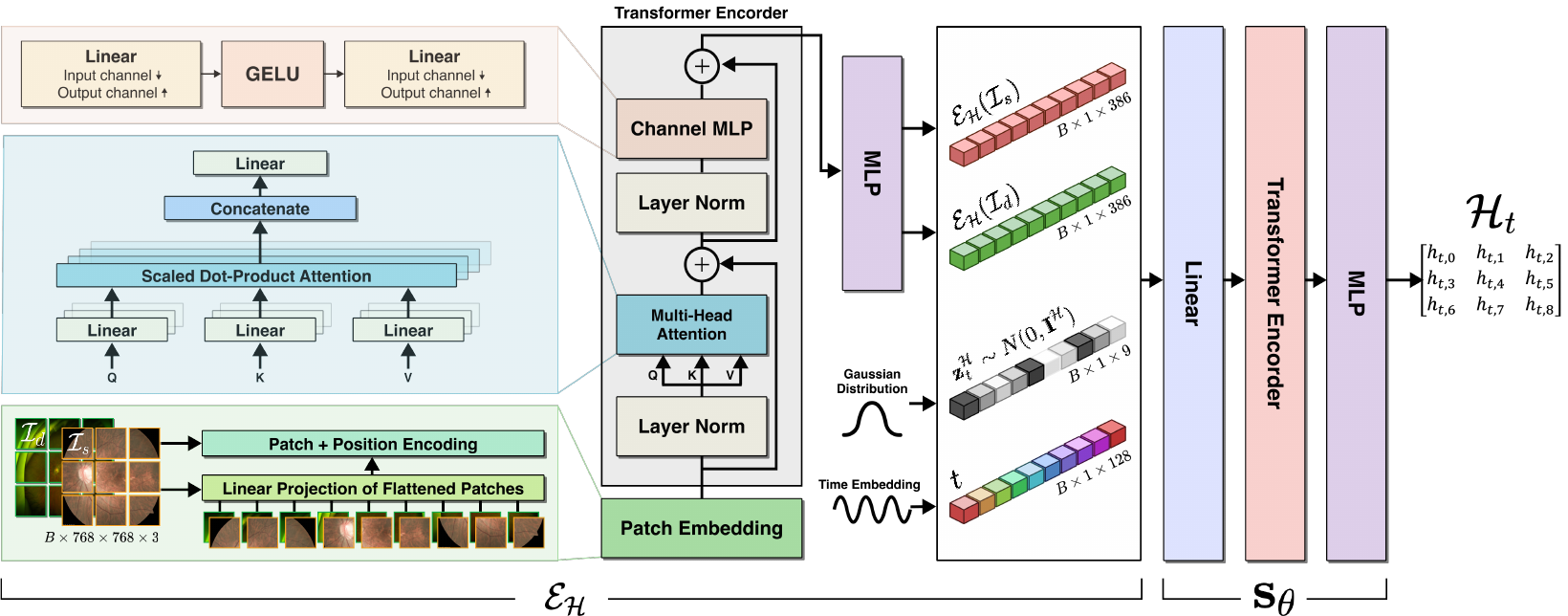}
    \caption{\textbf{Architectural details of the network components in the homography estimation path.} \(\mathcal{E}_\mathcal{H}\) and \(\mathbf{s}_{\boldsymbol{\theta}}\) first estimate the homography parameters \(\mathcal{H}_t\).}
    \label{fig:FIG_s_theta}
\end{figure*}

\begin{figure*}[h!]
    \centering
    \includegraphics[width=1.0\textwidth]{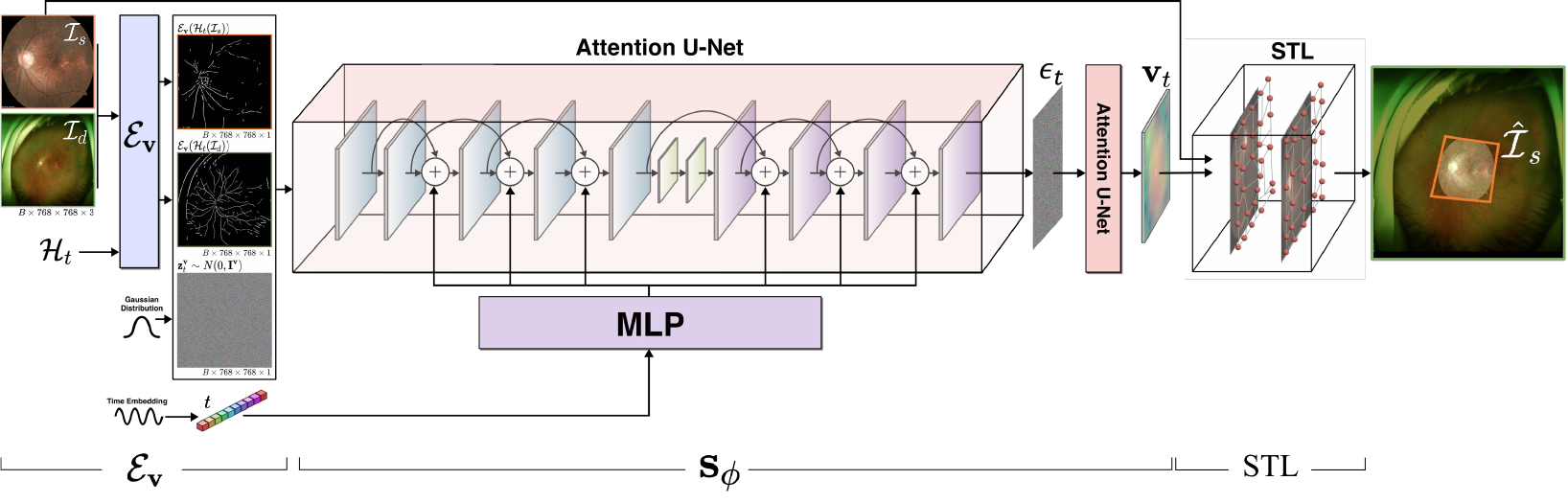}
    \caption{\textbf{Architectural details of the network components in the displacement field estimation path.} \(\mathcal{E}_{\mathbf{v}}\) and \(\mathbf{s}_{\boldsymbol{\phi}}\) then estimate the displacement field parameters \(\mathbf{v}_t\), while STL generates the warped image \(\hat{\mathcal{I}}_s\).}
    \label{fig:FIG_s_phi}
\end{figure*}

\begin{table}[t]
\centering
\caption{Summary of Notations Used in ADM}
\label{tab:notation}
\begin{tabular}{ll}
\toprule
\textbf{Symbol} & \textbf{Description} \\
\midrule
$\mathcal{I}_s$, $\mathcal{I}_d$ & Source and destination images (\textbf{Input}) \\
$\mathbf{x}_s$, $\mathbf{x}_d$ & Pixel grids of source and destination images \\
$\mathcal{H}$ & Homography parameters \\
$\mathbf{v}$ & Pixel-wise displacement field \\
$W(\cdot; \mathcal{H})$ & Grid warping function using homography \\
$\mathcal{E}_\mathcal{H}$, $\mathcal{E}_\mathbf{v}$ & Encoders for homography and displacement \\
$\mathbf{s}_\theta$, $\mathbf{s}_\phi$ & Score networks for $\mathcal{H}$ and $\mathbf{v}$ \\
$\mathcal{H}_t$, $\mathbf{v}_t$ & Noisy variables at timestep $t$ \\
$\mathbf{z}^{\mathcal{H}}_t$, $\mathbf{z}^{\mathbf{v}}_t$ & Gaussian noise at timestep $t$ \\
$\beta_t^{\mathcal{H}}$, $\beta_t^{\mathbf{v}}$ & Diffusion step sizes for $\mathcal{H}$ and $\mathbf{v}$ \\
$\bar{\alpha}_t^{\mathcal{H}}$, $\bar{\alpha}_t^{\mathbf{v}}$ & Cumulative noise schedule at $t$ \\
$t$ & Diffusion timestep \\
$\hat{\mathcal{I}}_s$ & Aligned image (\textbf{Output}) \\
STL & Spatial Transformer Layer \\
$\mathcal{L}_{\mathbf{s}}$, $\mathcal{L}_{\mathbf{x}}$, $\mathcal{L}_{\text{R}}$ & Score, pixel, and regularization losses \\
$g_L$ & Guidance weight for adaptive sampling \\
\bottomrule
\end{tabular}
\end{table}

\subsection{Active Diffusion Model for Image Alignment}
We denote the given source and destination image pair as \( \mathcal{I}_s\) and \(\mathcal{I}_d \), and the corresponding pixel grids \( {\mathbf x}_{s} = \{x_i | i \in (0, …, w \times h-1) \} \) and \( {\mathbf x}_{d}  = \{x_j | j \in (0, …, w \times h-1) \}\), where \( (w,h) \) denotes the width and height of each image.
We define the parameters for image alignment, namely homography and displacement vectors, as \({\mathcal{H}} = \left\{h_{0}, \cdots, h_{8}\right\}\) and as \( {\mathbf {v}} =  \{v_i | i \in (0, \cdots, w \times h-1) \}\), respectively.
The goal is to find \({\mathcal{\mathcal{H}}}\) and \( {\mathbf {v}}\) such that \( \mathcal{I}_s(W({x}_i;\mathcal{\mathcal{H}}) + v_{i})\) and \(\mathcal{I}_d(x_j) \) correspond to the same pixel locations, where \(W\) denotes the grid warping function.
If we treat \( \mathcal{\mathcal{H}} \) and \(\mathbf{v}\) as random variables, their conditional probability densities are defined as \(p(\mathcal{\mathcal{H}}|\mathcal{I}_s, \mathcal{I}_d)\) and \(p({\mathbf {v}}|\mathcal{I}_s, \mathcal{I}_d, \mathcal{\mathcal{H}})\), where $\mathbf {v}$ is conditional on $\mathcal{\mathcal{H}}$.

The overall structure of \textsc{ADM} is illustrated in Fig.~\ref{fig:FIG_ADM}. 
We set $\mathcal{\mathcal{H}}$ and $\mathbf{v}$ as variables for robust estimation using score-based models. 
The noise conditional neural network models $\mathbf{s}_{\boldsymbol\theta}(\mathcal{H}_{t},t\vert \mathcal{I}_s,\mathcal{I}_d)$ and $\mathbf{s}_{\boldsymbol\phi}({\mathbf v}_{t},t\vert \mathcal{I}_s,\mathcal{I}_d,\mathcal{H}_{t})$, with parameters ${\boldsymbol\theta}$ and ${\boldsymbol\phi}$, are trained to match the conditional score functions $\nabla_{\mathcal{\mathcal{H}}} \log p({\mathcal{\mathcal{H}}}_{t}|\mathcal{I}_s,\mathcal{I}_d)$ and $\nabla_{{\mathbf v}} \log p(\mathbf v_t|\mathcal{I}_s,\mathcal{I}_d,\mathcal{H}_{t})$, based on the denoising score~\cite{song2021scorebased,chan2024tutorial}.
These models enable iterative sampling, following Eq.~\ref{eq:ddpm_samp}, and are both conditioned on the input image 
pair $\mathcal{I}_s$ and $\mathcal{I}_d$ through the features computed from custom encoders $\mathcal{E}_\mathcal{H}$ and $\mathcal{E}_{\mathbf v}$.
Since $s_{\boldsymbol\phi}({\mathbf v}_{t},t\vert \mathcal{I}_s,\mathcal{I}_d,\mathcal{H}_{t})$ is conditioned on the output of $\mathbf{s}_{\boldsymbol\theta}(\mathcal{H}_{t},t\vert \mathcal{I}_s,\mathcal{I}_d)$, end-to-end training is required.

During inference, estimates \(\mathcal{H}_t\) and \(\mathbf{v}_t\) are iteratively updated by $\mathbf{s}_{\boldsymbol\theta}(\mathcal{H}_{t},t\vert \mathcal{I}_s,\mathcal{I}_d)$ and $\mathbf{s}_{\boldsymbol\phi}({\mathbf v}_{t},t\vert \mathcal{I}_s,\mathcal{I}_d,\mathcal{H}_{t})$.
As $\mathbf{s}_{\boldsymbol\phi}({\mathbf v}_{t},t\vert \mathcal{I}_s,\mathcal{I}_d,\mathcal{H}_{t})$ is conditioned on \(\mathcal{H}_t\), \(\mathcal{H}_t\) effectively serves as guidance for estimating \(\mathbf{v}_t\).
That is, the globally warped image from $\mathcal{I}_s$ using the estimated $\mathcal{H}_t$ is used together with $\mathcal{I}_d$ to estimate $\mathbf{v}_t$, as in~\cite{10821896}.
In addition, we add a guidance term during the inference of \(\mathcal{H}_t\), thereby interconnecting the estimation paths for $\mathcal{\mathcal{H}}$ and $\mathbf{v}$, as explained in more detail in Sec.~\ref{sec:sampling}.
The aligned image $\hat{\mathcal{I}}_s$ is generated from \(\mathcal{I}_s\), \(\mathcal{H}_t\), and $\mathbf{v_{t}}$ using the spatial transformer layers (STL), adapted from the spatial transformer network~\cite{jaderberg2015spatial}.

\begin{figure*}[ht]
    \centering
    \includegraphics[width=1.0\textwidth]{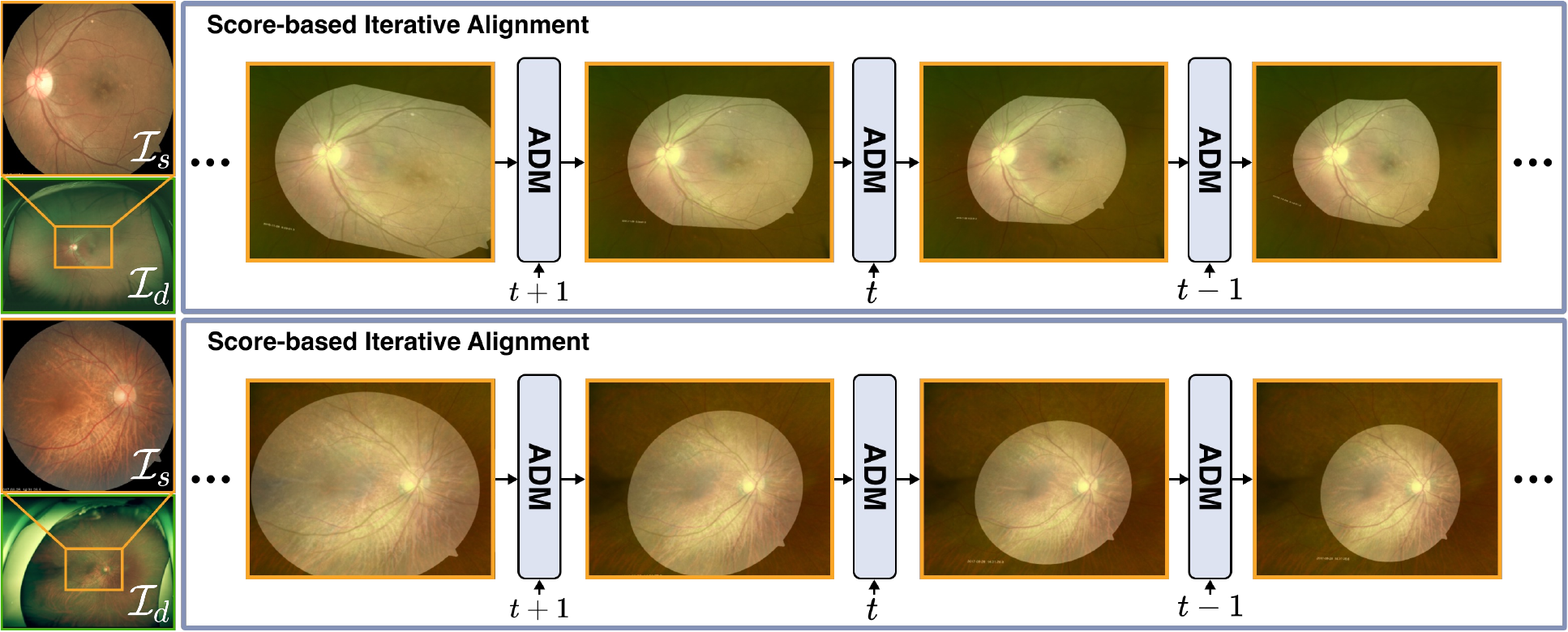}
    \caption{{\textbf{Score-based Iterative Alignment.}} \textsc{ADM} progressively predicts the global transform and local deformations to align SFI-UWFI pairs.}
    \label{fig:FIG_SAMP}
\end{figure*}

\subsection{Network Architectures} 
Here, we provide a detailed explanation of each component in \textsc{ADM}: \(\mathcal{E}_\mathcal{H}\), $\mathbf{s}_{\boldsymbol\theta}$, \(\mathcal{E}_\mathbf{v}\), $\mathbf{s}_{\boldsymbol\phi}$, and STL.
The symbols used for the components of \textsc{ADM} are summarized in Tab.~\ref{tab:notation}.

\subsubsection{Components in the Homography Estimation Path}
The detailed structure of \(\mathcal{E}_\mathcal{H}\) and $\mathbf{s}_{\boldsymbol\theta}$ is illustrated in Fig.~\ref{fig:FIG_s_theta}. 

\(\mathcal {E}_\mathcal{H}\) is based on a vision transformer model, initialized with the pre-trained DINO~\cite{caron2021emerging} model, and then fine-tuned on our dataset.
\(\mathcal {E}_\mathcal{H}\) takes an image of size \(768 \times 768 \times 3\) as input and outputs a \(384\)-dimensional feature vector.

For $\mathbf{s}_{\boldsymbol\theta}$, we use a combination of Linear + Transformer Encoder + MLP, chosen for its efficacy in parameter estimation for imaging tasks~\cite{wang2023posediffusion, liu2023geometrized}. 
Specifically, $\mathbf{s}_{\boldsymbol\theta}$ accepts \(384\)-dimensional feature embeddings \({\mathcal{E}_\mathcal{H}}(\mathcal{I}_s)\) and \({\mathcal{E}_\mathcal{H}}(\mathcal{I}_d)\), the noised \(9\)-dimensional homography $q({\mathcal{\mathcal{H}}}_{t}|{\mathcal{\mathcal{H}}}_{0}) \sim \mathcal{N}_{\mathcal{\mathcal{H}}}(\sqrt{\bar{\alpha}^{\mathcal{\mathcal{H}}}_t} {\mathcal{\mathcal{H}}}_0, (1 - \bar{\alpha}^{\mathcal{\mathcal{H}}}_t)\mathbf{I})$, and the \(128\)-dimensional vector for sampling step vector \(t\) with time embedding~\cite{dhariwal2021diffusion}. 
The linear layer maps the concatenated \(905\)-dimensional input to a \(512\)-dimensional primitive vector, which is then passed to the transformer encoder to generate an intermediate feature with $512$ dimensions.
This intermediate feature is finally interpreted to infer the \(9\)-dimensional homography parameters \(\mathcal{H}_t\), via the MLP layers.

\subsubsection{Components in the Displacement Field Estimation Path}
The detailed structure of the combination of \(\mathcal{E}_\mathbf{v}\), $\mathbf{s}_{\boldsymbol\phi}$, and STL is illustrated in Fig.~\ref{fig:FIG_s_phi}.

For \(\mathcal {E}_\mathbf{v}\), a vessel enhancement filter~\cite{BahadarKhan2016morph} is used to produce a simplified binary image.
\(\mathcal {E}_\mathbf{v}\) takes a \(768 \times 768 \times 3\) image as input and outputs a \(768 \times 768 \times 1\) image.

We use a U-net~\cite{ronneberger2015u} based network structure~\cite{dhariwal2021diffusion} for $\mathbf{s}_{\boldsymbol\phi}$.
In this structure, the latent feature has a spatial dimension scaled by \(\times 1/24\) and a channel dimension scaled by \(\times 128\), with an input image of size \(w \times h \times 1\). 
$\mathbf{s}_{\boldsymbol\phi}$ takes \(768 \times 768 \times 1\) dimensional feature embeddings \({\mathcal{E}_\mathbf{v}}(\mathcal{H}_{t}(\mathcal{I}_s))\) and \({\mathcal{E}_\mathbf{v}}(\mathcal{I}_d)\), a noisy image $q({\epsilon}_{t}|{\mathbf \epsilon}_{0}) \sim \mathcal{N}_{\mathbf \epsilon}(\sqrt{\bar{\alpha}^{\epsilon}_t} \mathbf{\epsilon}_0, (1 - \bar{\alpha}^{\epsilon}_t)\mathbf{I})$, and the aforementioned \(128\)-dimensional vector for sampling step \(t\) with time embedding~\cite{dhariwal2021diffusion}.
$\mathbf{s}_{\boldsymbol\phi}$ first estimates the \(768 \times 768 \times 1\) dimensional noise \(\epsilon_t\) and then calculates the score of the displacement \(\mathbf v_t\), which has \(768 \times 768 \times 2\) dimensions, using a U-net~\cite{ronneberger2015u} based structure from \cite{balakrishnan2018unsupervised}.

STL incorporates layers from the spatial transformer network~\cite{jaderberg2015spatial}, which samples new pixel values for the warped image \(\hat{\mathcal{I}}_s\) using interpolation between the globally transformed source image \(\mathcal{H}_{t}(\mathcal{I}_s)\) and the displacement field \(\mathbf{v_t}\).

\subsection{Training the Active Diffusion Model}
\label{sec:training}
The loss function \(\mathcal{L}\) for the end-to-end training of the \textsc{ADM} is defined as:
\begin{equation}\label{eq_full}
\begin{split}
\mathcal{L} = \mathcal{L}_{\mathbf{s}} + \lambda_{\mathbf{x}}\mathcal{L}_{\mathbf{x}} + \lambda_{\text{R}}\mathcal{L}_{\text{R}},
\end{split}
\end{equation}
where \(\mathcal{L}_{\mathbf{s}}\), \(\mathcal{L}_{\mathbf{x}}\), and \(\mathcal{L}_\text{R}\) denote the denoising score matching loss, pixel matching loss, and regularization loss, respectively. 
\(\lambda_{\mathbf{x}}\) and \(\lambda_{\text{R}}\) control the relative importance of each terms.
Each component, along with additional training details, will be described in the following subsections.

\subsubsection{Score Matching Loss}
Score-based Markov chain equations for $\mathcal{\mathcal{H}}$ and $\mathbf v$ are defined as follows:
\begin{equation}
    {\mathcal{\mathcal{H}}}_{t-1} = \frac{1}{\sqrt{1-\beta^{\mathcal{\mathcal{H}}}_{t}}}{\mathcal{\mathcal{H}}}_{t} + 
    \beta^{\mathcal{\mathcal{H}}}_{t} \mathbf{s}_{\boldsymbol{\theta}^{\mathbf{*}}}(\mathcal{\mathcal{H}}, t) + \sqrt{\beta^{\mathcal{\mathcal{H}}}_{t}} \mathbf{z}^{\mathcal{\mathcal{H}}}_t,
\label{eq:ddpm_samp_H}
\end{equation}
\begin{equation}
    \mathbf{v}_{t-1} = \frac{1}{\sqrt{1-\beta^{\mathbf{v}}_{t}}}
    \mathbf{v}_{t} + \beta^{\mathbf{v}}_{t} \mathbf{s}_{\boldsymbol{\phi}^{\mathbf{*}}}(\mathbf{v}, t) + \sqrt{\beta^{\mathbf{v}}_{t}} \mathbf{z}^{\mathbf{v}}_t.
    \label{eq:ddpm_samp_v}
\end{equation}
Here, $\mathbf{z}^{\mathcal{H}}_t \sim \mathcal{N}(\mathbf{0}, \mathbf{I^{\mathcal{H}}})$
and $\mathbf{z}^{\mathbf v}_t \sim \mathcal{N}(\mathbf{0}, \mathbf{I^{\mathbf v}})$ are the standard noise terms for $\mathcal{H}$ and $\mathbf{v}$, respectively.

The loss functions for score matching with $\mathbf{s}_{\boldsymbol\theta}(\mathcal{H}_{t},t)$ and $\mathbf{s}_{\boldsymbol\phi}({\mathbf v}_{t},t)$ are defined as:
\begin{equation}
\resizebox{.9\hsize}{!}{$
    \mathcal{L}_{\mathbf{s}_{\boldsymbol{\theta}}}
    =
    \mathbb{E}_{p{(\mathcal{H})}} 
    \left[\left\lVert \mathbf{s}_{\boldsymbol{\theta}}\left(\mathcal{H}_{0}+{\sqrt{1 - \bar{\alpha}^{\mathcal{H}}_t}}\mathbf{z}^{\mathcal{H}}_t \;\middle|\; \mathcal{I}_s,\mathcal{I}_d\right) + \frac{\mathbf{z}^{\mathcal{H}}_t}{\sqrt{1 - \bar{\alpha}^{\mathcal{H}}_t}} \right\rVert_2^2\right]
    $},
    \label{eq:dsm_h}
\end{equation}
\begin{equation}
\resizebox{.9\hsize}{!}{$
    \mathcal{L}_{\mathbf{s}_{\boldsymbol{\phi}}}    
    = 
    \mathbb{E}_{p{({\mathbf v})}} 
    \left[\left\lVert \mathbf{s}_{\boldsymbol{\phi}}\left({\mathbf v}_{0}+{\sqrt{1 - \bar{\alpha}^{{\mathbf v}}_t}}\mathbf{z}^{{\mathbf v}}_t \;\middle|\; \mathcal{I}_s,\mathcal{I}_d, \mathcal{H}_{t}\right) + \frac{\mathbf{z}^{{\mathbf v}}_t}{\sqrt{1 - \bar{\alpha}^{{\mathbf v}}_t}} \right\rVert_2^2\right]
    $},
    \label{eq:dsm_v}
\end{equation}
derived from the Gaussian noise kernels $p_{\alpha^{\mathcal{H}}_{t}} ({\mathcal{H}_{t}}|\mathcal{H}_{0}) = \mathcal{N}(\mathcal{H}_{t};\sqrt{\alpha^{\mathcal{H}}_{t}}\mathcal{H}_{0}, (1-\alpha^{\mathcal{H}}_{t})\mathbf{I})$ and $p_{\alpha^{\mathbf{v}}_{t}} ({\mathbf{v}_{t}}|\mathbf{v}_{0}) = \mathcal{N}(\mathbf{v}_{t};\sqrt{\alpha^{\mathbf{v}}_{t}}\mathbf{v}_{0}, (1-\alpha^{\mathbf{v}}_{t})\mathbf{I})$, respectively. 
Note that the expectation notation is simplified here and omits explicit sampling of noise and timesteps for clarity.

The combined loss function for score matching is defined as the the weighted sum of the two individual losses:
\begin{equation}
\mathcal{L}_{\mathbf{s}} = \mathcal{L}_{\mathbf{s}_{\boldsymbol{\theta}}} + \delta_{\mathbf{s}} \mathcal{L}_{\mathbf{s}_{\boldsymbol{\phi}}},
\label{eq:loss_score}
\end{equation}
where $\delta_{\mathbf{s}}$ is a weight coefficient that is dynamically scheduled to suppress the influence of potentially inaccurate homography parameter estimation during the early stages of training.

\subsubsection{Pixel Matching Loss} 
Since Eq.~\ref{eq:dsm_h} directly measures the squared error difference between two homography matrices, it may fail to reflect the actual pixel-wise displacement induced by these transformations.
We incorporate the \emph{p-norm} measure, proposed by Je and Park~\cite{je2015pnorm}, which defines a metric between two homography matrices, based on the source image points $x_{i} \in \mathbf{x}_{s}$ for which the homography is applied, as follows:
\begin{equation}\label{p_norm}
\mathcal{L}^{\mathcal{H}}_{\mathbf{x}}\left( \mathcal{H}_{t}, \mathcal{H}_{0} \right) = \sum_{{x}_{i} \in {\mathbf x}_{s}}\left( {\lVert \mathcal{H}_{t}{x}_{i}-\mathcal{H}_{0}{x}_{i} \rVert^{p}} \right)^{1/p}.
\end{equation}

To further encourage appearance consistency after alignment, we additionally define a pixel matching loss \(\mathcal{L}^{a}_{\mathbf{x}}\), following ~\cite{kim2022diffusemorph}:
\begin{equation}
\mathcal{L}^{a}_{\mathbf{x}} = -NCC({\mathcal{E}}_{\mathbf{v}}(\hat{\mathcal{I}}_s),{\mathcal{E}}_{\mathbf{v}}(\mathcal{I}_d)),
\label{L_L}
\end{equation}
where \(NCC\) represents the normalized cross-correlation between the aligned pixel appearances.

The combined loss function for pixel matching is defined as the sum of the two terms above:
\begin{equation}
\mathcal{L}_{\mathbf{x}} = \delta_{\mathbf{x}} \mathcal{L}^{\mathcal{H}}_{\mathbf{x}} + \mathcal{L}^{a}_{\mathbf{x}},
\label{eq:loss_pixel}
\end{equation}
where $\delta_{\mathbf{x}}$ is a time-dependent weight, defined as a quadratic function that increases over time, starting from zero at \(t = T\).

\subsubsection{Regularization Loss}
The regularization loss function is defined as follows:
\begin{multline}
    \mathcal{L}_{\text{R}}
     = \delta_{\text{R}} \mathcal{L}^{\mathcal{H}}_{\text{R}} + \mathcal{L}^{\mathbf{v}}_{\text{R}} \\
    = \delta_{\text{R}} \sum_{{x}_{i} \in {\mathbf x}_{s}}\left( {\lVert \mathcal{H}_{t}{x}_{i}-{x}_{i} \rVert^{p}} \right)^{1/p} + \sum{ \lVert \nabla \mathbf{v} \rVert^{2}},
\label{eq:loss_reg}
\end{multline}
where $\delta_{\text{R}}$ is a time-dependent weight, defined as a quadratic function that decreases over time, reaching zero at \(t = 0\).
$\mathcal{L}^{\mathcal{H}}_{\text{R}}$ is equivalent to $\mathcal{L}^{\mathcal{H}}_{\mathbf{x}}\left( \mathcal{H}_{t}, \mathbf{I} \right)$, and it penalizes deviation of the estimate $\mathcal{H}_{t}$ from deviating too far from the identity mapping.
$\mathcal{L}^{\mathbf{v}}_{\text{R}}$ constrains the estimate $\mathbf{v}_{t}$ to avoid large discontinuities in the displacement vectors.

\begin{algorithm}[t]
\caption{ADM Inference Algorithm}
\label{alg:adm_sampling}
\KwIn{Source image $\mathcal{I}_s$, Destination image $\mathcal{I}_d$, \\
\hspace{2.75em} Initial noise $\mathcal{H}_T$, $\mathbf{v}_T$, Step size $\beta_t^\mathcal{H}$, $\beta_t^\mathbf{v}$, \\
\hspace{2.75em} Random noise $\mathbf{z}^{\mathcal{H}}_t \sim \mathcal{N}(\mathbf{0}, \mathbf{I^{\mathcal{H}}})$, $\mathbf{z}^{\mathbf v}_t \sim \mathcal{N}(\mathbf{0}, \mathbf{I^{\mathbf v}})$}
\KwOut{Aligned image $\hat{\mathcal{I}}_s$}

\For{$t = T$ \KwTo $1$}{
    \textbf{// --- Homography Update ---} \\
    Predict score: $\mathbf{s}_\theta(\mathcal{H}_t, t \mid \mathcal{I}_s, \mathcal{I}_d)$ \\
    Warp source image: $\mathcal{H}_t(\mathcal{I}_s)$ \

    Estimate $\mathbf{v}_t \leftarrow \mathbf{s}_\phi(\mathcal{H}_t(\mathcal{I}_s), \mathcal{I}_d, t)$ \\
    $\hat{\mathcal{I}}_s \leftarrow \text{STL}(\mathcal{I}_s, \mathcal{H}_t, \mathbf{v}_t)$ \\
    Compute appearance loss: $\mathcal{L}^a_{\mathbf{x}} \leftarrow -\text{NCC}(\hat{\mathcal{I}}_s, \mathcal{I}_d)$ \;
    Compute guidance: $\nabla_{\mathcal{H}_t} \mathcal{L}^a_{\mathbf{x}}$ \\
    Apply guidance: $\hat{\mathbf{s}}_\theta \leftarrow \mathbf{s}_\theta - g_L \cdot \nabla_{\mathcal{H}_t} \mathcal{L}^a_{\mathbf{x}}$ \\
    Update: $\mathcal{H}_{t-1} \leftarrow \frac{1}{\sqrt{1 - \beta_t^\mathcal{H}}} \mathcal{H}_t + \beta_t^\mathcal{H} \cdot \hat{\mathbf{s}}_\theta + \sqrt{\beta_t^\mathcal{H}} \cdot \mathbf{z}_t^\mathcal{H}$ \\

    \textbf{// --- Displacement Update ---} \\
    Update: $\mathbf{v}_{t-1} \leftarrow \frac{1}{\sqrt{1 - \beta_t^\mathbf{v}}} \mathbf{v}_t + \beta_t^\mathbf{v} \cdot \mathbf{s}_\phi + \sqrt{\beta_t^\mathbf{v}} \cdot \mathbf{z}_t^\mathbf{v}$ \\
}

Final output: $\hat{\mathcal{I}}_s \leftarrow \text{STL}(\mathcal{I}_s, \mathcal{H}_0, \mathbf{v}_0)$ \;

\end{algorithm}

\subsection{Inference Strategies}\label{sec:inference}
Fig.~\ref{fig:FIG_SAMP} illustrates the gradual alignment of \(\mathcal{I}_s\) and \(\mathcal{I}_d\) through \textsc{ADM}. 
Here, we aim to explain the specific details of the sampling process in \textsc{ADM}.
\subsubsection{Input Adaptive Guided Sampling}
\label{sec:sampling}
Guided sampling, denoted by the blue arrow of the \textsc{ADM} components shown in Fig.~\ref{fig:FIG_ADM}, allows parameter estimation to be further adapted to the input image pair.
Among the loss function terms described in Sec.~\ref{sec:training}, the term \(\mathcal{L}^{a}_{\mathbf{x}}\) directly depends on the input images \(\mathcal{I}_d\) and \(\hat{\mathcal{I}}_s\), which are warped using the estimates $\mathcal{H}_t$ and \(\mathbf{v}_t\).
The gradient of this term with respect to the parameters guides the parameter optimization to adapt to the given input.

In each sampling step, we adjust the predicted \(\mathbf{s}_{\boldsymbol{\theta}}({\mathcal{H}_{t}},t | \mathcal{I}_s, \mathcal{I}_d)\) by the gradient of \(L^{a}_{\mathbf{x}}\) as follows:
\begin{equation}\label{eq_guidance}
\begin{split}
\hat{\mathbf{s}}_{\boldsymbol{\theta}}({\mathcal{H}_{t}},t | \mathcal{I}_s, \mathcal{I}_d) = \mathbf{s}_{\boldsymbol{\theta}}({\mathcal{H}_{t}},t | \mathcal{I}_s, \mathcal{I}_d) - g_{L} \, \nabla_{\mathcal{H}_t} \mathcal{L}^{a}_{\mathbf{x}},
\end{split}
\end{equation}
where \(g_{L}\) controls the strength of guidance.
That is, an initial \({\mathcal{\mathcal{H}}}_{t}\) is computed from the homography estimation path and provided to the displacement field estimation path, after which the derivative from the displacement field estimation path on \({\mathcal{\mathcal{H}}}_t\) is used to compute the modified homography parameters, at each timestep.
We note that empirical observations led us to apply the guidance only to \({\mathcal{\mathcal{H}}}_{t}\) and not to \(\mathbf{v}_{t}\), as applying the guidance to both parameters may result in contradictory effects.
This process is described in Algorithm~\ref{alg:adm_sampling}.

\subsubsection{Iterative \textsc{ADM}} 
We apply \textsc{ADM} iteratively using the output \(\hat{\mathcal{I}}_s\) as the new input \(\mathcal{I}_s\) to achieve better results.
Since \(\hat{\mathcal{I}}_s\) is more closely aligned with \(\mathcal{I}_d\) than \(\mathcal{I}_s\), we expect improved results with just a few additional iterations.

\begin{figure*}[h!] 
    \centering
    \includegraphics[width=1.0\textwidth]{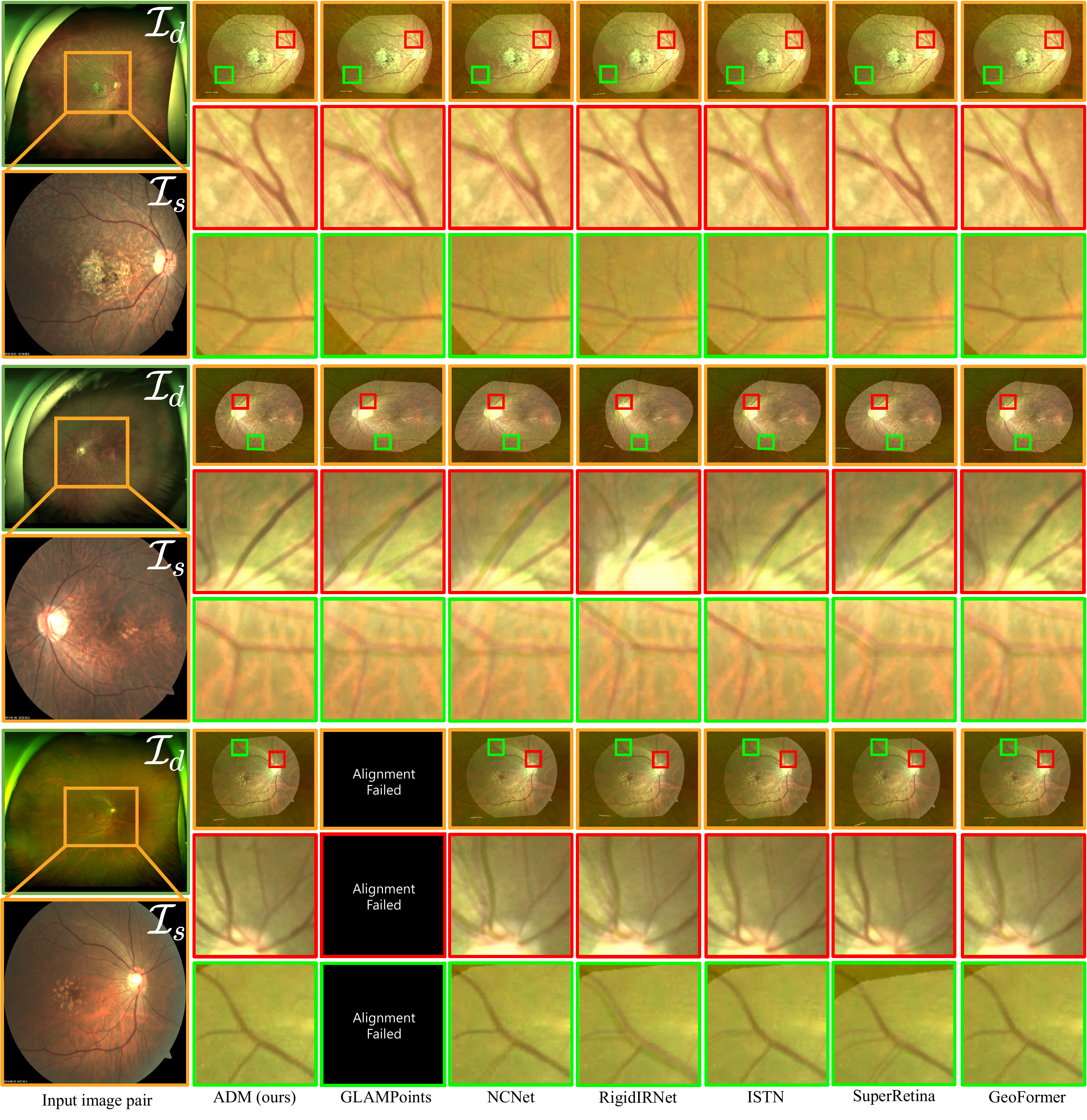}
    \caption{\textbf{Qualitative comparisons of direct homography estimation methods using sample images from the KBSMC dataset.} We illustrate alignment results for SFI-UWFI pairs with GLAMPoints~\cite{truong2019glampoints}, NCNet~\cite{rocco2020ncnet}, RigidIRNet~\cite{de2019deep}, ISTN~\cite{lee2019istn}, SuperRetina~\cite{liu2022semi}, GeoFormer~\cite{liu2023geometrized}, and \textsc{ADM} (ours).}
    \label{fig:FIG_KBSMC_direct}
\end{figure*}

\section{Experiments}\label{sec:exp}
\subsection{Datasets}
We evaluated our algorithm using a dataset from the Kangbuk Samsung Medical Center (KBSMC) Ophthalmology Department, which includes $3744$ SFIs and paired but non-aligned UWFIs, collected between 2017 and 2019\footnote{This study adhered to the tenets of the Declaration of Helsinki and was approved by the Institutional Review Boards (IRB) of Kangbuk Samsung Hospital (No. KBSMC 2019-08-031). The study is a retrospective review of medical records, and the data were fully anonymized prior to processing. The IRB waived the requirement for informed consent.}. 
The SFIs in this dataset exhibit an approximate \(\times 1\) to \(\times 4\) difference in scale compared to the UWFIs, which were captured from the same patients.

We randomly split the dataset into a training set ($3370$ pairs) and an evaluation set ($374$ pairs). 
SFIs are resized to \(768 \times 768\) pixels, and UWFIs are resized and cropped accordingly to match the image resolutions. 
For cropping, we apply random positions to augment the SFI-UWFI pairs. 
Pseudo-ground-truth homography matrices for aligning SFIs and UWFIs were generated through manual keypoint annotations.

Furthermore, we evaluated the proposed method on the public FIRE dataset~\cite{hernandezmatas2017fire}, which includes $134$ pairs of images with the corresponding ground truth homography matrices.

\subsection{Baselines for Comparison}
We compare our \textsc{ADM} with several baselines using SFI-UWFI pairs from the KBSMC dataset. The compared methods include SIFT~\cite{lowe2004distinctive} (with RANSAC~\cite{fischler1987ransac}), SuperPoint~\cite{detone2018superpoint}, GLAMpoints~\cite{truong2019glampoints}, NCNet~\cite{rocco2020ncnet}, RigidIRNet~\cite{de2019deep}, ISTN~\cite{lee2019istn}, SuperRetina~\cite{liu2022semi}, GeoFormer~\cite{liu2023geometrized}, DLKFM~\cite{zhao2021deep}, and MCNet~\cite{zhu2024mcnet}. 
These baselines are trained from scratch on our dataset.

For comparisons on the FIRE dataset~\cite{hernandezmatas2017fire}, we provide results for six additional methods: SuperGlue~\cite{sarlin2020superglue}, R2D2~\cite{revaud2019r2d2}, REMPE~\cite{hernandez2020rempe}, DKM~\cite{edstedt2023dkm}, LoFTR~\cite{sun2021loftr}, and ASpanFormer~\cite{chen2022aspanformer}. 
For these methods, we reprint the values reported in \cite{liu2023geometrized}.

\subsection{Evaluation Metrics}
To assess the performance of \textsc{ADM}, we employ the approach of CEM~\cite{charles2003dual} for measuring the median error (MEE) and the maximum error (MAE), following conventions from related works~\cite{wang2015robust, liu2022semi, truong2019glampoints, liu2023geometrized}. 
The success of the alignment results (Success Rate) is categorized as:
\begin{itemize}
    \item \textit{Failed} (no homography created),
    \item \textit{Acceptable} (MAE \(<\) 50 and MEE \(<\) 20),
    \item \textit{Inaccurate} (otherwise).
\end{itemize}

We also measure the Area Under the Curve (AUC)~\cite{hernandezmatas2017fire}, which computes the expectation of the \textit{acceptable} rate with respect to the error of 25, as described in~\cite{liu2022semi}. 
Additionally, we calculate the mean AUC (mAUC) in Tables~\ref{tab:private} and~\ref{tab:public}, which represents the mean value of the AUC over the total number of image pairs.

\subsection{Implementation Details}
We used the AdamW~\cite{loshchilov2017decoupled} optimizer with a learning rate of \(0.001\), \(\beta_1 = 0.9\), \(\beta_2 = 0.999\), and \(\epsilon = 10^{-8}\) to train \textsc{ADM}. 
The weight decay was applied every \(100\)K iterations with a decay rate of \(0.01\). 
The learning rate was halved every \(150K\) iterations. 
We used a batch size of $3$ and trained the model for more than \(3 \times 10^{7}\) iterations on an NVIDIA RTX 4090 GPU. 
Images of size \(768 \times 768\) were fed into the network to train both the homography estimation path and the displacement field estimation path simultaneously. 
Data augmentation was performed by applying random rotations of \(90^\circ\), \(180^\circ\), or \(270^\circ\) to the images. 
The limited choices of rotation were chosen to maintain the structure of the retinal images resulting from the acquisition protocols. 
On a single NVIDIA RTX 4090 GPU, inference took an average of \(47.12\) seconds and consumed \(1.2\)GB of memory for a \(768 \times 768\) image pair. 

The timestep index \(t\) was sampled in the range \(0 < t < T\), with \(T\) set to \(100\) for $\mathbf{s}_{\boldsymbol{\theta}}$ and \(500\) for $\mathbf{s}_{\boldsymbol{\phi}}$. 
The coefficients \(\lambda_{\mathbf{x}}\) and \(\lambda_{\text{R}}\) were set to \(1\) and \(0.1\), respectively. 
The coefficient \(\delta_{\mathbf{s}}\), which adjusts the degree of loss depending on $\mathbf{s}_{\boldsymbol{\phi}}$ relative to $\mathbf{s}_{\boldsymbol{\theta}}$, was set to \(0\) for the first quarter of the training steps and \(1\) thereafter. 
The coefficients \(\delta_{\mathbf{x}}\) and \(\delta_{\text{R}}\) were set to \((T-t)^2/T^2\) and \(10^{-3} \times t^2/T^2\), respectively. 
Additionally, for \(\mathcal{L}^\mathcal{H}_{\mathbf{x}}\) and \(\mathcal{L}^\mathcal{H}_{\text{R}}\), we sampled \(20\) image points and set \(p\) to \(2\).

\begin{figure*}[thb]
    \centering
    \includegraphics[width=1.0\textwidth]{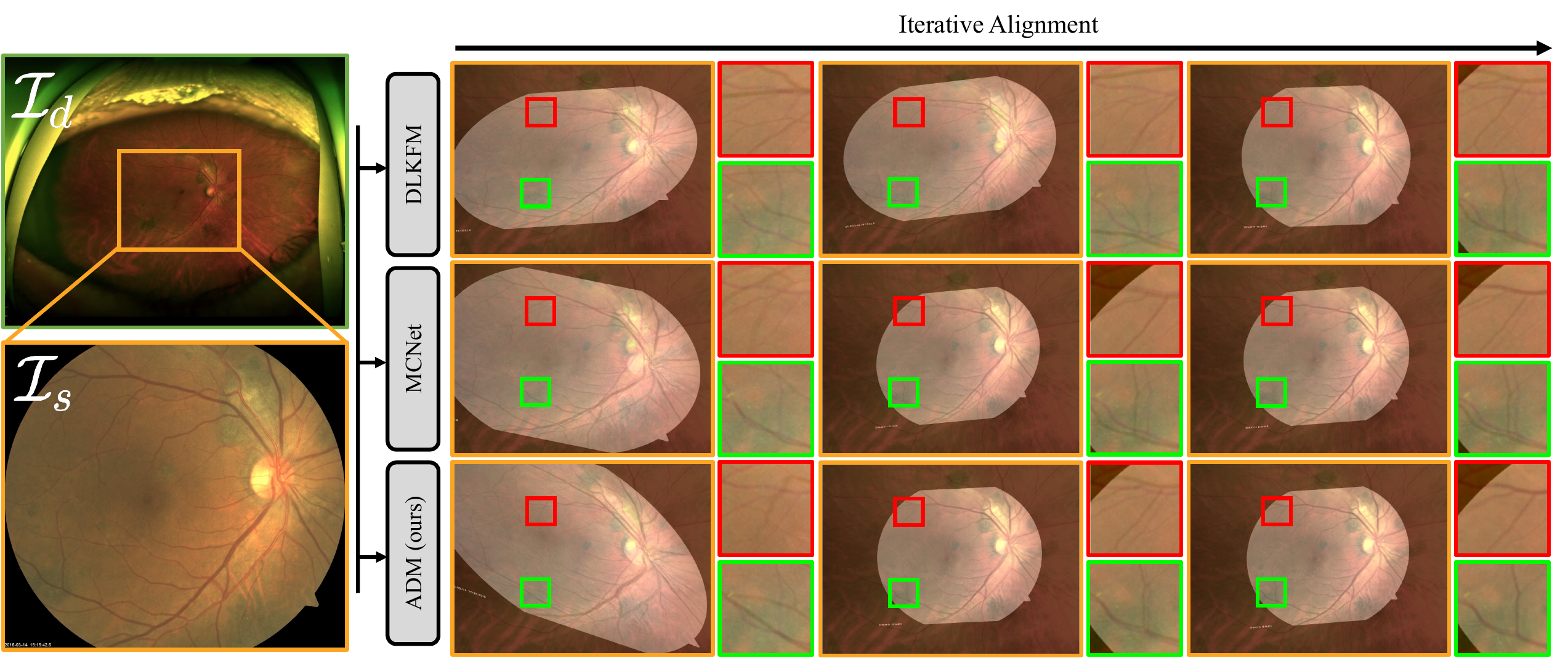}
    \caption{\textbf{Qualitative comparisons of iterative homography estimation methods on sample images from the KBSMC dataset.} Alignment results between SFI-UWFI pairs are illustrated using DLKFM~\cite{zhao2021deep}, MCNet~\cite{zhu2024mcnet}, and \textsc{ADM} (ours).}
    \label{fig:FIG_KBSMC_iterative}
\end{figure*}

\subsection{Comparative Evaluation on SFI-UWFI Pairs Dataset}
Quantitative comparisons of our private KBSMC dataset are presented in Tab.~\ref{tab:private}.
As UWFIs may lack distinctive regions compared to SFIs, alignment results using keypoints detected by SIFT~\cite{lowe2004distinctive} were suboptimal. 
Self-supervised keypoint detection methods such as SuperPoint~\cite{detone2018superpoint} and GLAMPoints~\cite{truong2019glampoints} faced challenges due to the significant domain gap between SFIs and UWFIs, resulting in difficulties in forming keypoint pairs and, consequently, exhibiting lower performance. 
On the other hand, methods such as NCNet~\cite{rocco2020ncnet} and GeoFormer~\cite{liu2023geometrized}, which use two images as input to find a suitable match, demonstrated relatively high performance.
Since the SuperRetina~\cite{liu2022semi} method is trained on annotated keypoints, we trained the model with varying numbers of sampled keypoints ($50$, $100$, $200$ pairs) generated based on pseudo-ground truth homography. 
Although performance increased with the number of training keypoints, it remained lower than that of GeoFormer.
The highest acceptable rate and mAUC benchmark values were achieved by \textsc{ADM}, with a 5.88\% point increase in the acceptable rate and a 5.2 point increase in mAUC, compared to the second-best method, GeoFormer, demonstrating the effectiveness of \textsc{ADM}.

\begin{table}[thb]
\renewcommand*{\arraystretch}{1.1}
\caption {\textbf{Comparative evaluation of the KBSMC dataset.}}
\label{tab:private}
\resizebox{1.0\columnwidth}{!}{
{\LARGE
\begin{tabular}{lcccc}
\hline
{\multirow{2}{*}{\textbf{Methods}}} & \multicolumn{3}{c}{\textbf{Success Rate (\%)}}                   & \multirow{2}{*}{\textbf{mAUC}} \\ \cline{2-4}
& \textit{Failed}   & \textit{Acceptable}   & \textit{Inaccurate}  &                       \\ \hline
SIFT~\cite{lowe2004distinctive}                 & 0        & 8.29           & 91.71          & 5.2                     \\
SuperPoint~\cite{detone2018superpoint}          & 0        & 9.09           & 90.91          & 8.7                     \\
GLAMpoints~\cite{truong2019glampoints}          & 0        & 9.89           & 90.11          & 8.4                    \\
NCNet~\cite{rocco2020ncnet}                     & 0       & 12.30           & 87.70          & 9.6                    \\
RigidIRNet~\cite{de2019deep}                    & 0       & 12.57  & 87.43  & 10.6                    \\
ISTN~\cite{lee2019istn}                         & 0       & 20.86  & 79.14  & 12.1                    \\
SuperRetina:50~\cite{liu2022semi}                & 0       & 15.78  & 84.22  & 10.1                    \\
SuperRetina:100~\cite{liu2022semi}               & 0       & 24.87  & 75.13  & 15.9                    \\
SuperRetina:200~\cite{liu2022semi}                & 0       & 34.76  & 65.24  & 22.3                    \\
GeoFormer~\cite{liu2023geometrized}             & 0       & \underline{36.10}           & \underline{63.90}          & \underline{24.1}                    \\ 

DLKFM~\cite{zhao2021deep}             & 0       & 22.73           & 77.27          & 13.5                    \\
MCNet~\cite{zhu2024mcnet}             & 0       & 32.89           & 67.11          & 20.9                    \\ \hline
\textsc{ADM} (ours)       &  0       & \bf{41.98}           & \bf{58.02}          & \bf{29.3}                    \\ 
\hline
\end{tabular}}}
\scriptsize{
The bold and underlined values denote the best and second-best results, respectively.\\
SuperRetina:X denotes the method trained with X manually annotated keypoints.}
\end{table}

Fig.~\ref{fig:FIG_KBSMC_direct} presents qualitative comparisons with direct homography estimation methods (GLAMPoints~\cite{truong2019glampoints}, NCNet~\cite{rocco2020ncnet}, RigidIRNet~\cite{de2019deep}, ISTN~\cite{lee2019istn}, SuperRetina~\cite{liu2022semi}, and GeoFormer~\cite{liu2023geometrized}) in Tab.~\ref{tab:private}. 
We exclude SIFT\cite{lowe2004distinctive} and SuperPoint~\cite{detone2018superpoint} from the qualitative results, as their alignment attempts mostly failed to produce meaningful transformations in our challenging setting.
The results of SuperRetina~\cite{liu2022semi} are obtained from training with annotations of $200$ keypoint pairs. 
We indicate the aligned area of SFIs overlaid on UWFIs with an orange box and provide comparisons by highlighting the overlaid warped images from each method in the top rows.
Additionally, we provide further comparisons in zoomed-in local regions, indicated by red and green boxes in the second and third rows.
Upon examination of the alignment through the overlaid images, it is evident that \textsc{ADM} provides the best alignment.

Fig.~\ref{fig:FIG_KBSMC_iterative} presents qualitative comparisons with iterative homography refinement methods (DLKFM~\cite{zhao2021deep} and MCNet~\cite{zhu2024mcnet}) in Tab.~\ref{tab:private}. 
Again, the orange box indicates aligned regions and the red and green boxes indicate zoomed-in regions, respectively.
The results according to the iterative optimization process are specified in every one-third of the total iteration steps of each work.
We adopted the basic optimization steps assumed in these works. 
Here, \textsc{ADM} is observed to converge slightly faster, with the most accurate final alignment.

\begin{figure}[thb]
    \centering
    \includegraphics[width=1.0\columnwidth]{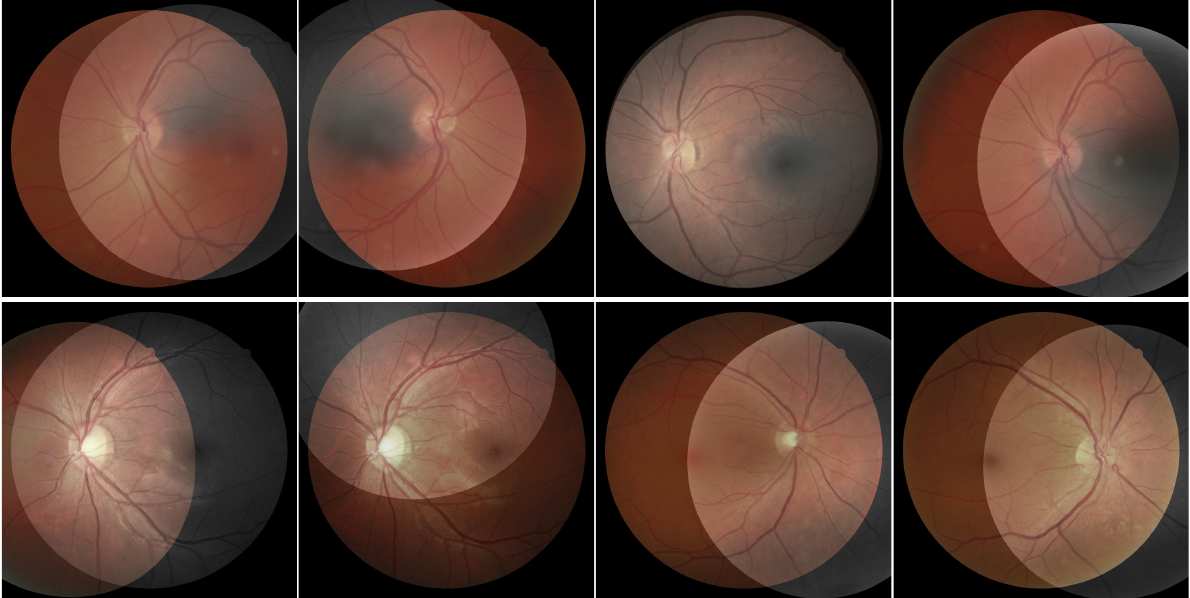}
    \caption{\textbf{Qualitative evaluation of \textsc{ADM} on the FIRE~\cite{hernandezmatas2017fire} dataset.} 
    }
    \label{fig:FIG_FIRE}
\end{figure}

\begin{table}[h]
\renewcommand*{\arraystretch}{1.1}
\caption {\textbf{Comparative evaluation of the FIRE~\cite{hernandezmatas2017fire} dataset.}}
\label{tab:public}
\resizebox{1.0\columnwidth}{!}{
{\LARGE
\begin{tabular}{lcccc}
\hline
{\multirow{2}{*}{\textbf{Methods}}} & \multicolumn{3}{c}{\textbf{Success Rate (\%)}}                   & \multirow{2}{*}{\textbf{mAUC}} \\ \cline{2-4}
& \textit{Failed}   & \textit{Acceptable}   & \textit{Inaccurate}  &                       \\ \hline
SIFT~\cite{lowe2004distinctive}                 & 0        & 79.85        & 20.15       & 57.3                     \\
SuperPoint~\cite{detone2018superpoint}          & 0        & 94.78        & 5.22        & 67.4                     \\
GLAMpoints~\cite{truong2019glampoints}          & 0        & 92.54        & 7.46        & 61.1                    \\
NCNet~\cite{rocco2020ncnet}                     & 0        & 85.82        & 14.18       & 61.2                    \\
SuperRetina~\cite{liu2022semi}                  & 0        & \bf{98.51}        & \bf{1.49}        & 75.5                    \\
GeoFormer~\cite{liu2023geometrized}             & 0        & \bf{98.51}        & \bf{1.49}      & \underline{75.6}                    \\ 
SuperGlue~\cite{sarlin2020superglue}            & 0.75     & 95.52        & 3.73        & 68.7                      \\
R2D2~\cite{revaud2019r2d2}                      & 0        & 95.52        & 4.48        & 71.1                  \\
REMPE~\cite{hernandez2020rempe}                 & 0        & \underline{97.01}        & \underline{2.99}        & 72.0                  \\
DKM~\cite{edstedt2023dkm}                       & 0        & 75.94        & 24.06       & 58.0                  \\
LoFTR~\cite{sun2021loftr}                       & 0        & 96.99        & 3.01        & 66.3                  \\
ASPanFormer~\cite{chen2022aspanformer}          & 0        & 91.73        & 8.27        & 70.6                  \\ \hline

\textsc{ADM} (ours)     & 0        & \bf{98.51}        & \bf{1.49}        & \bf{76.0}                    \\

\hline
\end{tabular}}}
\scriptsize{
The bold and underline values denote the best and second best results, respectively.\\
All comparative evaluation results except \textsc{ADM} are reproduced from \cite{liu2023geometrized}. \\
Among the 134 pairs, P\_37 was labeled as \textit{Inaccurate} due to an annotation error.
}
\end{table}

\subsection{Comparative Evaluation on SFI-SFI pairs Dataset} 
Quantitative comparative evaluation of \textsc{ADM} on the $134$ image pairs in the FIRE~\cite{hernandezmatas2017fire} dataset is presented in Tab.~\ref{tab:public}. 
It is observed that \textsc{ADM} achieves the highest benchmark performance in terms of both \textit{Acceptable} Rate and mAUC metric, albeit slightly.
The margin of improvement compared to existing methods was smaller than in the KBSMC dataset, as the difficulty of alignment was considerably lower.
Some examples of the alignment results of \textsc{ADM} are shown in Fig.~\ref{fig:FIG_FIRE}. 

To facilitate effective alignment of SFI-SFI pairs within the FIRE~\cite{hernandezmatas2017fire} dataset, we employed a self-supervised learning approach for training \textsc{ADM}. 
Specifically, we utilized only the SFIs from the KBSMC dataset to synthetically create random homography matrices and their corresponding paired translated images in real-time. These generated image pairs and the corresponding homography matrices were used to train \textsc{ADM}, which was then fine-tuned with the FIRE~\cite{hernandezmatas2017fire} dataset.
We note that comparison methods SuperRetina~\cite{liu2022semi} and GeoFormer~\cite{liu2023geometrized} also mention similar pre-training processes.

\begin{table}[thb]
{
\renewcommand*{\arraystretch}{1.0}
\caption {\textbf{Ablative evaluation on inference strategy.}}
\label{tab:ablation_guidance}
\resizebox{1.0\columnwidth}{!}{
{\LARGE
\begin{tabular}{clllccccc}
\hline
\multicolumn{3}{c}{\multirow{2}{*}{\textbf{Methods}}}     &  & \multicolumn{4}{c}{\textbf{KBSMC} \boldmath$(n=374)$}       &  \\ \cline{5-8}
\multicolumn{3}{c}{}                              &  & \textit{Failed} & \textit{Acceptable} & \textit{Inaccurate} & mAUC &  \\ \hline
\multicolumn{2}{c}{\textsc{ADM} : \textit{full}}                 &                                       &  &    0    &     41.98       &     58.02      &   29.3   &  \\
\multicolumn{1}{r}{}         & \multicolumn{2}{l}{without Iterative \textsc{ADM}}                          &  &    0    &     39.84       &     60.16      &   27.8   &  \\
\multicolumn{1}{r}{}         & \multicolumn{2}{l}{without \(\nabla_{\mathcal{H}_{t}} \mathcal{L}^{a}_{\mathbf{x}}\) guidance}     &  &    0    &     31.82       &     68.18      &   17.1   &  \\
 \hline
\multicolumn{3}{c}{\multirow{2}{*}{\textbf{Methods}}}     &  & & \multicolumn{4}{c}{\textbf{FIRE}~\cite{hernandezmatas2017fire} \boldmath$(n=134)$}        \\ \cline{5-8} 
\multicolumn{3}{c}{}                              &  & \textit{Failed} & \textit{Acceptable} & \textit{Inaccurate} & mAUC \\ \hline
        \multicolumn{2}{c}{\textsc{ADM} : \textit{full}}                 &                                      &  &     0   &      98.51      &   1.49         &   76.0   \\
        \multicolumn{1}{r}{}         & \multicolumn{2}{l}{without Iterative \textsc{ADM}}                         &  &     0   &      97.76      &   2.24         &   74.8   \\
\multicolumn{1}{r}{}         & \multicolumn{2}{l}{without \(\nabla_{\mathcal{H}_{t}} \mathcal{L}^{a}_{\mathbf{x}}\) guidance}            &  &     0   &      94.77      &   5.23         &   71.8   \\
 \hline
\end{tabular}
}
}
}

\end{table}

\begin{table}[thb]
{
\renewcommand*{\arraystretch}{1.0}
\caption {\textbf{Ablative evaluation on dynamic scheduling.}}
\label{tab:ablation_scheduling}
\resizebox{1.0\columnwidth}{!}{
{\scriptsize
\begin{tabular}{ccccccccc}
\hline
\multirow{2}{*}{$\delta_{\mathbf{s}}$} & \multirow{2}{*}{\(\delta_{\mathbf{x}}\)} & \multirow{2}{*}{\(\delta_{\text{R}}\)}    &  & \multicolumn{4}{c}{\textbf{KBSMC} \boldmath$(n=374)$}       &  \\ \cline{5-8}
\multicolumn{3}{c}{}                              &  & \textit{Failed} & \textit{Acceptable} & \textit{Inaccurate} & mAUC &  \\ \hline
\ding{51}        &     \ding{51}           &    \ding{51}                                  &  &     0    &     41.98       &     58.02      &   29.3   &  \\
\ding{51}       &      \ding{55}          &     \ding{55}                                 &  &     0    &     37.43       &     62.57      &   26.6   &  \\
\ding{55}       &    \ding{51}            &     \ding{55}                                 &  &     0    &     34.60       &     65.40      &   20.9   &  \\
\ding{55}       &    \ding{55}            &     \ding{51}                                 &  &     0    &     36.63       &     63.37      &   21.1   &  \\
\ding{55}       &    \ding{55}            &     \ding{55}                                 &  &     0    &     34.22       &     65.78      &   20.5   &  \\
 \hline
\multirow{2}{*}{$\delta_{\mathbf{s}}$} & \multirow{2}{*}{\(\delta_{\mathbf{x}}\)} & \multirow{2}{*}{\(\delta_{\text{R}}\)}   & & \multicolumn{4}{c}{\textbf{FIRE}~\cite{hernandezmatas2017fire} \boldmath$(n=134)$}  \\ \cline{5-8} 
\multicolumn{3}{c}{}                              &  & \textit{Failed} & \textit{Acceptable} & \textit{Inaccurate} & mAUC \\ \hline
\ding{51}        &     \ding{51}           &    \ding{51}                                  &  &    0   &      98.51      &   1.49         &   76.0   \\
\ding{51}       &      \ding{55}          &     \ding{55}                                 &  &     0   &      98.51      &   1.49         &   75.8   \\
\ding{55}       &    \ding{51}            &     \ding{55}                                 &  &     0   &      95.52      &   4.48         &   73.2   \\
\ding{55}       &    \ding{55}            &     \ding{51}                                 &  &     0    &     96.27       &  3.73     &   73.3   &  \\
\ding{55}       &    \ding{55}            &     \ding{55}                                 &  &     0    &     94.78       &  5.22     &   71.9   &  \\
 \hline
\end{tabular}
}
}
}

\end{table}

\begin{table}[thb]
{
\renewcommand*{\arraystretch}{1.0}
\caption {\textbf{Ablative evaluation on network architecture.}}
\label{tab:ablation_archtecture}
\resizebox{1.0\columnwidth}{!}{
{\LARGE
\begin{tabular}{clllccccc}
\hline
\multicolumn{3}{c}{\multirow{2}{*}{\textbf{Methods}}}     &  & \multicolumn{4}{c}{\textbf{KBSMC} \boldmath$(n=374)$}       &  \\ \cline{5-8}
\multicolumn{3}{c}{}                              &  & \textit{Failed} & \textit{Acceptable} & \textit{Inaccurate} & mAUC &  \\ \hline
\multicolumn{3}{l}{Transformer $\mathbf{s}_{\boldsymbol{\theta}}$ + CNN $\mathbf{s}_{\boldsymbol{\phi}}$} &&    0    &     41.98       &     58.02      &   29.3   &  \\
\multicolumn{3}{l}{Transformer $\mathbf{s}_{\boldsymbol{\theta}}$ + Transformer $\mathbf{s}_{\boldsymbol{\phi}}$} &&    0    &     38.77       &     69.23      &   28.5   &  \\
\multicolumn{3}{l}{CNN $\mathbf{s}_{\boldsymbol{\theta}}$ + CNN $\mathbf{s}_{\boldsymbol{\phi}}$} &&    0    &     34.49       &     65.61      &   21.4   &  \\
\multicolumn{3}{l}{CNN $\mathbf{s}_{\boldsymbol{\theta}}$ + Transformer $\mathbf{s}_{\boldsymbol{\phi}}$} &&    0    &     31.55       &     68.45      &   18.5   &  \\
 \hline
\multicolumn{3}{c}{\multirow{2}{*}{\textbf{Methods}}}     &  & & \multicolumn{4}{c}{\textbf{FIRE}~\cite{hernandezmatas2017fire} \boldmath$(n=134)$}        \\ \cline{5-8} 
\multicolumn{3}{c}{}                              &  & \textit{Failed} & \textit{Acceptable} & \textit{Inaccurate} & mAUC \\ \hline
\multicolumn{3}{l}{Transformer $\mathbf{s}_{\boldsymbol{\theta}}$ + CNN $\mathbf{s}_{\boldsymbol{\phi}}$} &&    0    &     98.51       &     1.49      &   76.0   &  \\
\multicolumn{3}{l}{Transformer $\mathbf{s}_{\boldsymbol{\theta}}$ + Transformer $\mathbf{s}_{\boldsymbol{\phi}}$} &&    0    &     98.51       &     1.49      &   75.6   &  \\
\multicolumn{3}{l}{CNN $\mathbf{s}_{\boldsymbol{\theta}}$ + CNN $\mathbf{s}_{\boldsymbol{\phi}}$} &&    0    &     97.01       &     2.99      &   72.5   &  \\
\multicolumn{3}{l}{CNN $\mathbf{s}_{\boldsymbol{\theta}}$ + Transformer $\mathbf{s}_{\boldsymbol{\phi}}$} &&    0    &     96.99       &     3.01      &   70.1   &  \\
 \hline
\end{tabular}}}
}

\end{table}

\begin{table}[thb]
{
\renewcommand*{\arraystretch}{1.0}
\caption {\textbf{Ablative evaluation of the test sample ratio on the KBSMC dataset.}}
\label{tab:ablation_testsamples}
\resizebox{1.0\columnwidth}{!}{
{\scriptsize
\begin{tabular}{cccccc}
\hline
\textbf{Percentage of train/test samples} & \textbf{90/10}   & \textbf{80/20}   & \textbf{70/30}   & \textbf{60/40}   & \textbf{50/50}   \\ \hline
mAUC               & 29.3 & 28.9 & 26.5 & 22.7 & 16.3 \\ \hline
\end{tabular}}}
}

\end{table}

\begin{figure}[thb]
    \centering
    \includegraphics[width=1.0\columnwidth]{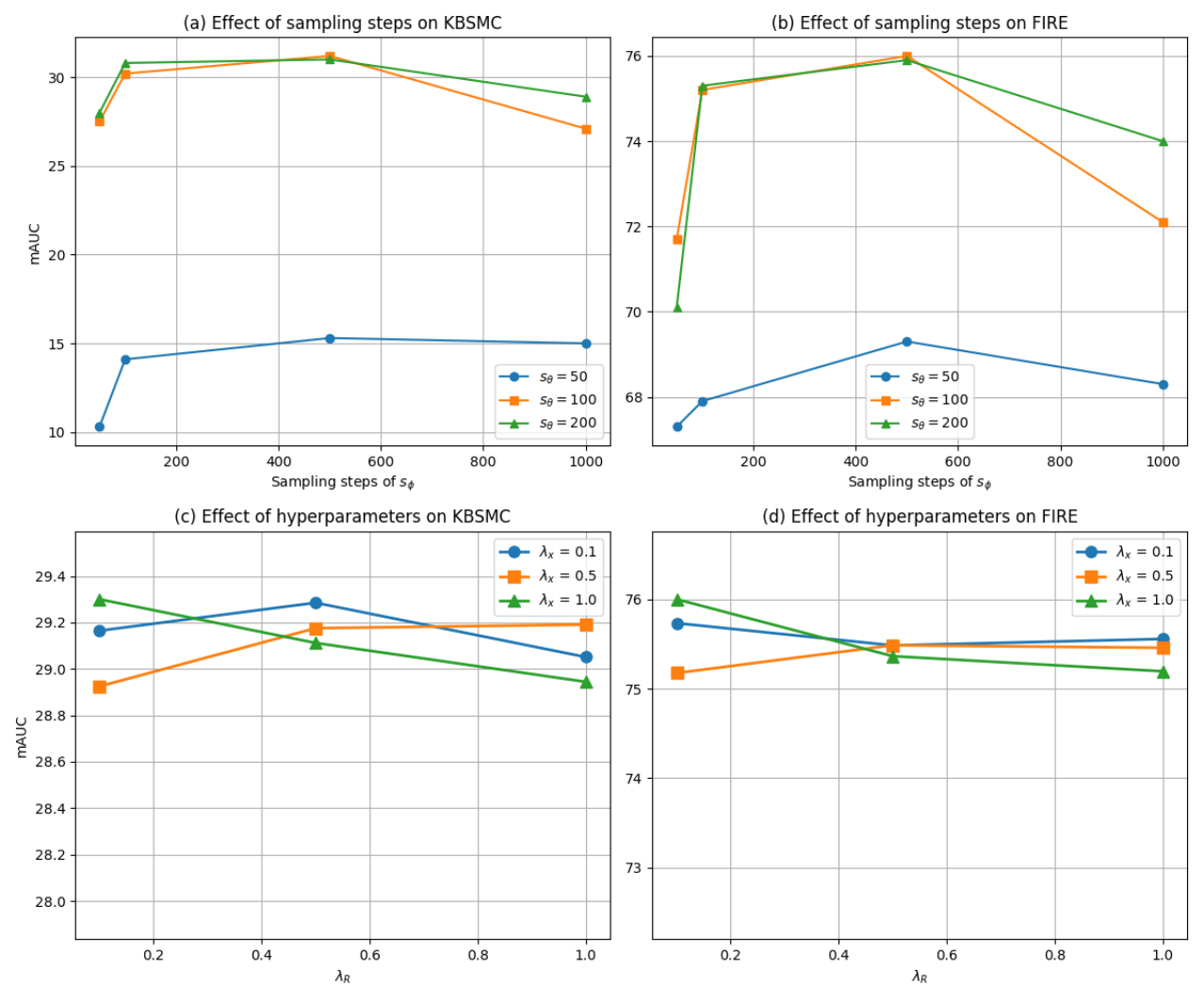}
    \caption{\textbf{Abalative evaluation of \textsc{ADM} per different sampling steps and hyperparameters.} 
    }
    \label{fig:FIG_ADDITER}
\end{figure}

\begin{table}[thb]
{
\renewcommand*{\arraystretch}{1.0}
\caption {\textbf{Ablative evaluation on degradations.}}
\label{tab:ablation_degradation}
\resizebox{1.0\columnwidth}{!}{
{\scriptsize
\begin{tabular}{llcc}
\hline
\multicolumn{2}{c}{\multirow{2}{*}{\textbf{Degradations}}} & \multicolumn{2}{c}{\textbf{mAUC}} \\ \cline{3-4} 
\multicolumn{2}{l}{}                              & \textbf{KBSMC}       & \textbf{FIRE}~\cite{hernandezmatas2017fire}       \\ \hline
\multirow{3}{*}{Gaussian noise}   & $\sigma=5$      & 28.9  & 75.7 \\
                                  & $\sigma=10$     & 26.1  & 72.6 \\
                                  & $\sigma=25$     & 22.5  & 68.2 \\ \hline
\multirow{3}{*}{Gaussian blur}    & $\sigma=1$      & 29.0  & 75.9 \\
                                  & $\sigma=2.5$    & 27.8  & 73.8 \\
                                  & $\sigma=5$      & 24.9  & 69.0 \\ \hline
\multirow{3}{*}{Low illumination} & $\alpha=0.75$ & 28.7  & 75.4 \\
                                  & $\alpha=0.5$  & 26.3  & 72.1 \\
                                  & $\alpha=0.25$ & 23.0   & 67.3 \\ \hline
\end{tabular}}
}
}
\end{table}

\subsection{Ablative Study}
To evaluate the impact of the inference strategy and components of our \textsc{ADM} on its performance, we performed ablative evaluations as follows.

\paragraph*{Inference Strategy}
This includes \textsc{ADM} variants without iterative refinement and without input-adaptive guided sampling.  
The results in Tab.~\ref{tab:ablation_guidance} demonstrate that iterative refinement of \(\mathcal{I}_s\) enhances performance, particularly for severely deformed image pairs in the KBSMC dataset.  
Furthermore, omitting guidance from the displacement field estimation path during homography estimation leads to a notable performance degradation, underscoring the importance of \textsc{ADM}'s dual diffusion structure and its guided sampling strategy.

\paragraph*{Dynamic Scheduling}
We conducted an ablation study to validate the effectiveness of each component in our dynamic scheduling strategy: \(\delta_{\mathbf{s}}\), \(\delta_{\mathbf{x}}\), and \(\delta_{\text{R}}\).  
As summarized in Table~\ref{tab:ablation_scheduling}, removing each element leads to performance degradation in both datasets.  
In particular, excluding \(\delta_{\mathbf{s}}\) caused the most significant drop in performance, underscoring its critical role in stabilizing homography estimation during the later sampling steps.  
The other components, \(\delta_{\mathbf{x}}\) and \(\delta_{\text{R}}\), also contributed consistently, supporting the effectiveness of our design in improving convergence and reliability during both training and inference.

\paragraph*{Network Architecture}
Our \textsc{ADM} adopts a Transformer-based architecture for \(\mathbf{s}_{\boldsymbol{\theta}}\) and a CNN-based architecture for \(\mathbf{s}_{\boldsymbol{\phi}}\), as described in the main text.  
This design choice is motivated by the fact that the Transformer, which excels at capturing global context, is well-suited for estimating global transformations such as homography, whereas the CNN, known for its ability to extract local features, is effective at modeling local deformations such as displacement fields~\cite{wu2021cvtintroducingconvolutionsvision}.  
However, we also explore variants that reverse the architectural assignments, applying a CNN-based architecture~\cite{detone2016deep} to \(\mathbf{s}_{\boldsymbol{\theta}}\) and a Transformer-based architecture~\cite{peebles2023scalablediffusionmodelstransformers} to \(\mathbf{s}_{\boldsymbol{\phi}}\).  
The results of this configuration are shown in Tab.~\ref{tab:ablation_archtecture}, supporting our hypothesis, as the empirical evidence aligns well with our architectural insights.

\paragraph*{Dataset Partitioning}
We currently use 10\% of the KBSMC dataset as the test set.  
As shown in Tab.~\ref{tab:ablation_testsamples}, we further evaluate the mAUC by progressively increasing the test set ratio, which accordingly reduces the proportion of training data.  
This analysis reveals a consistent decline in performance as the amount of training data decreases.

\paragraph*{Sampling Steps}
We conduct an ablation study to examine the effect of the number of sampling iterations for the global transformation estimator $\mathbf{s}_{\boldsymbol{\theta}}$ and the local deformation estimator $\mathbf{s}_{\boldsymbol{\phi}}$ on the final performance. 
As shown in Fig.~\ref{fig:FIG_ADDITER}~(a) and~(b), increasing the number of steps for $\mathbf{s}_{\boldsymbol{\theta}}$ leads to a substantial improvement in mAUC, indicating that accurate global transformation estimation requires a sufficient number of iterations. 
Notably, the performance gain saturates beyond 100 steps, suggesting a point of diminishing returns. 
In contrast, increasing the number of iterations for $\mathbf{s}_{\boldsymbol{\phi}}$ yields only a modest improvement up to 500 steps, after which the performance begins to degrade. 
We hypothesize that this drop results from the characteristics of 2D image-level diffusion models, where excessive iterations may introduce artifacts or over-smooth features, thereby impairing alignment~\cite{aithal2024understandinghallucinationsdiffusionmodels}. 
These findings suggest that while global estimation benefits from a greater number of iterations, local refinement must be carefully balanced to avoid over-processing.

\paragraph*{Hyperparameters}
As shown in Equation~\ref{eq_full}, our loss function comprises a primary score-matching term and two auxiliary terms, each weighted by a corresponding hyperparameter. 
While the auxiliary losses assist in guiding the optimization process, they play a secondary role. 
As reported in Fig.~\ref{fig:FIG_ADDITER}~(c) and~(d), the overall performance remains stable across a wide range of values for \(\mathcal{L}_{\mathbf{x}}\) and \(\mathcal{L}_\text{R}\), indicating that our method is robust to the choice of these hyperparameters.

\paragraph*{Degraded Inputs}
To evaluate the robustness of our method, we simulate three common types of image degradation frequently used in vision research: Gaussian noise, Gaussian blur, and low illumination. 
Gaussian noise is introduced by adding zero-mean white noise to the image, with the noise level controlled by the standard deviation parameter \(\sigma\)~\cite{zhang2017beyond}. 
Gaussian blur is applied via a smoothing filter, where \(\sigma\) determines the spread of the blur kernel~\cite{ledig2017photo}. 
Low illumination is simulated by scaling the pixel intensities by a factor \(\alpha \in (0,1)\), with smaller \(\alpha\) values producing darker images~\cite{chen2018learning}. 
These synthetic corruptions are widely used to assess the robustness of vision models~\cite{hendrycks2019benchmarking}. 
As shown in Tab.~\ref{tab:ablation_degradation}, our model maintains stable performance across all degradation types, despite being trained solely on clean images.

\begin{figure}[t!]
    \centering
    \includegraphics[width=1.0\columnwidth]{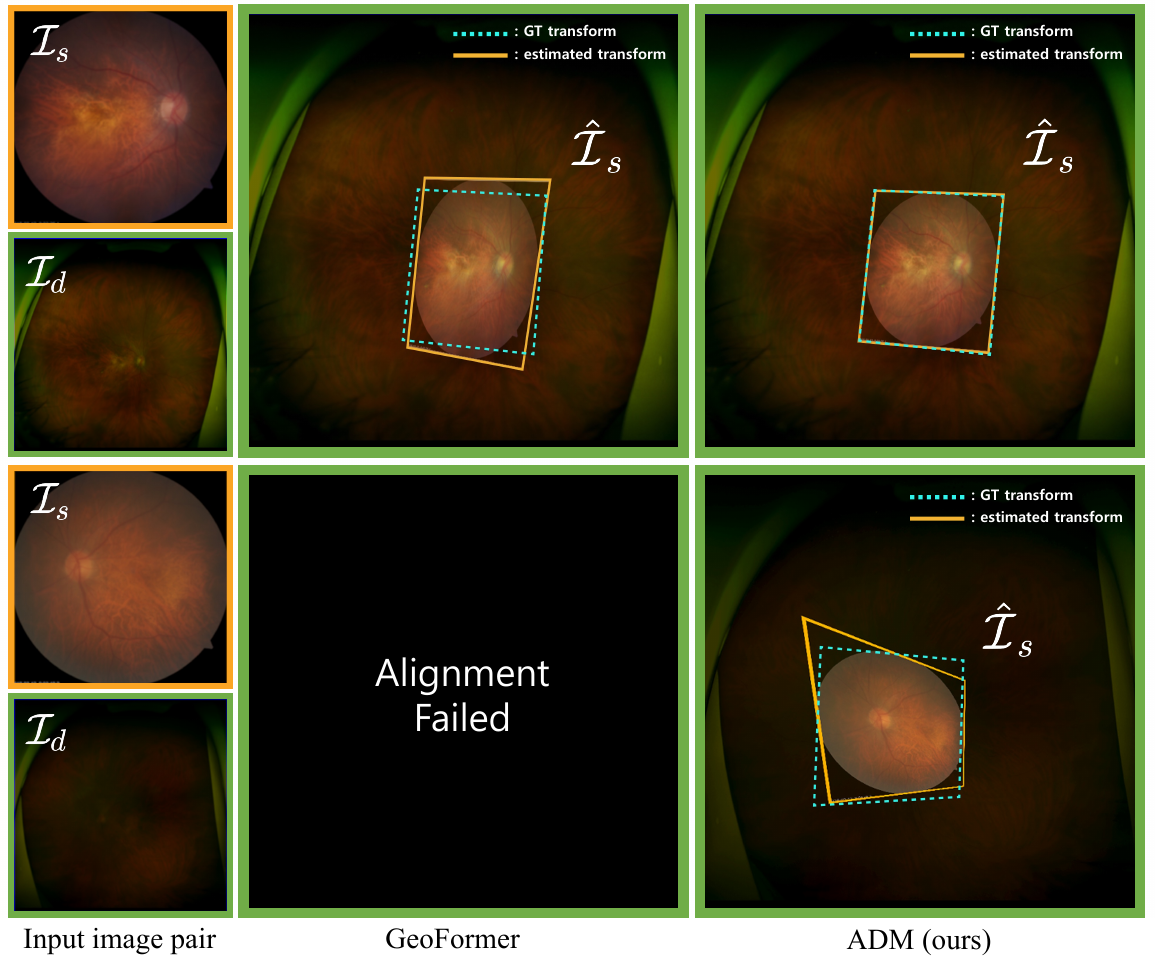}
    \caption{\textbf{Failure cases.} The transformation estimation results of \textsc{ADM} and the baseline GeoFormer~\cite{liu2023geometrized} are presented on two highly challenging registration samples from the KBSMC dataset.
    }
    \label{fig:FIG_FAIL}
\end{figure}

\section{Discussion}
In the following, we discuss several key aspects that warrant consideration in relation to the proposed \textsc{ADM}.

As shown in Fig.~\ref{fig:FIG_FAIL}, our \textsc{ADM} accurately estimates the transformation for moderately challenging SFI-UWFI pairs, where GeoFormer~\cite{liu2023geometrized} fails. 
However, in more extreme cases, particularly when vessel structures become indistinct due to strong blur or low illumination, $\mathbf{s}_{\boldsymbol{\phi}}$ struggles to estimate local deformations reliably. 
This failure is largely attributable to the decreased visibility of vascular features, which are essential for effective displacement estimation. 
Although our method utilizes a vessel enhancement filter~\cite{BahadarKhan2016morph} in the preprocessing stage, its performance is limited under severe degradations, as the filter is applied uniformly regardless of image quality.
In practice, such degraded UWFIs are frequently encountered, underscoring a critical limitation that warrants further investigation.
Enhancing robustness through adaptive preprocessing or dynamic weighting in the displacement path could mitigate such issues. 
While our method demonstrates strong overall performance, reducing the domain gap between SFIs and UWFIs and improving resilience to severe degradation remain important directions for future research, especially in medical applications requiring reliable registration under suboptimal imaging conditions.

Another important issue is the high inference time (47.12 seconds) associated with the iterative estimation process of $\mathbf{s}_{\boldsymbol{\phi}}$, which is substantially longer than that of key baselines such as SuperRetina (2.5 seconds) and GeoFormer (1.5 seconds). 
Although recent one-step denoising diffusion models~\cite{song2023consistency,salimans2022progressive} present a promising avenue for accelerating inference, their limited accuracy and adaptability to fine-grained tasks like registration remain significant challenges. 
Alternatively, strategies such as knowledge distillation~\cite{meng2023distillationguideddiffusionmodels} or selective iteration pruning of $\mathbf{s}_{\boldsymbol{\phi}}$ could potentially reduce runtime while preserving alignment quality. 
These observations highlight the need for future research to focus on enhancing robustness to image degradation and reducing inference time without compromising alignment accuracy, especially in time-sensitive clinical applications.

While this iterative process increases inference time, it also offers a significant advantage over discrete and feedforward models such as GeoFormer~\cite{liu2023geometrized}. 
Since \textsc{ADM} employs score-based Langevin dynamics, it is capable of progressively refining alignment estimates during inference without relying solely on training-time representations. 
This characteristic enables greater adaptability to previously unseen image pairs, especially in cases exhibiting substantial appearance variations between SFIs and UWFIs. 
Although the overall numerical improvements may appear modest, our observations indicate that gains are concentrated in more challenging cases, such as those with severe degradation or extensive lesion areas where local structures are less distinct. 
We anticipate that incorporating degradation-aware modeling in future work will further enhance the performance of \textsc{ADM}. 
Given the increasing clinical adoption of UWFI and the scarcity of prior research specifically targeting cross-modal alignment in this domain, we consider our approach a meaningful and timely contribution.

The structural nature of medical images allows \textsc{ADM} to generalize beyond SFI-UWFI data. 
Since $\mathbf{s}_{\boldsymbol{\phi}}$ estimates local displacement fields based on anatomical structures like vessels, the method is applicable across various imaging modalities and clinical environments where such structures are preserved.

Lastly, our \textsc{ADM} incorporates a regularization loss during training to mitigate failures in warped image generation caused by incorrect or divergent homography predictions from $\mathbf{s}_{\boldsymbol{\theta}}$ within the iterative global transformation estimation process. 
Nevertheless, the KBSMC dataset poses a considerable challenge for registration, and in some instances, the resulting warped images exhibit severe distortion or complete misalignment. 
This issue is regarded as a primary factor contributing to the observed performance degradation on the KBSMC dataset. 
To address this limitation, a more robust iterative procedure could be developed by detecting and discarding unreliable global transformation predictions during intermediate iterations, followed by their re-estimation. 
Such a strategy is anticipated to enhance alignment accuracy, particularly for highly challenging datasets such as KBSMC.

\section{Conclusion}
In this paper, we propose a novel cross-modal image alignment method, \textsc{ADM}. 
By employing score-matched diffusion models as dynamic components within a Langevin Markov chain for stochastic iterative estimation, we demonstrate that \textsc{ADM} achieves robust alignment results on the extremely challenging task of aligning SFI-UWFI pairs. 
We introduce several customized components, including \emph{p-norm} regularization during training, input-adaptive guided sampling, and an iterative inference scheme for \textsc{ADM}. 
A comparative evaluation against recent state-of-the-art methods shows that \textsc{ADM} outperforms competing approaches, despite a moderate increase in sampling time attributable to its dual diffusion model architecture. 
This trade-off between accuracy and computational cost has practical implications, particularly in applications where robustness is of paramount importance. 
We believe the \textsc{ADM} framework holds strong potential, especially for advancing methods aimed at UWFI enhancement.

\bibliographystyle{unsrt}
\bibliography{main}

\end{document}